\documentclass[10pt,letterpaper,english,final]{article}
\newif\ifdraft
\draftfalse

\usepackage[top=0.85in,left=2.75in,footskip=0.75in]{geometry}
\usepackage{amsmath,amssymb}
\usepackage{changepage}
\usepackage[utf8x]{inputenc}
\usepackage{textcomp,marvosym}
\usepackage{cite}
\usepackage{nameref,hyperref}
\usepackage[right]{lineno}
\usepackage{microtype}
\DisableLigatures[f]{encoding = *, family = * }
\usepackage[table]{xcolor}
\usepackage{array}
\newcolumntype{+}{!{\vrule width 2pt}}
\newlength\savedwidth

\ifdraft
\usepackage{setspace}
\fi
\raggedright
\usepackage[aboveskip=1pt,labelfont=bf,labelsep=period,justification=raggedright,singlelinecheck=off]{caption}

\makeatletter
\renewcommand{\@biblabel}[1]{\quad#1.}
\makeatother
\usepackage{lastpage,fancyhdr,graphicx}
\usepackage{epstopdf}
\fancyheadoffset[L]{2.25in}
\fancyfootoffset[L]{2.25in}
\usepackage{babel}
\addto\captionsenglish{} 
\usepackage[subrefformat=simple,labelformat=simple,labelsep=colon]{subcaption}

\usepackage{graphicx}
\usepackage{rotating}
\usepackage{multirow}
\usepackage{theorem}
\usepackage[ruled,lined]{algorithm2e}
\usepackage{paralist}
\usepackage{verbatim}
\usepackage{balance}
\usepackage[multiple]{footmisc}
\usepackage{placeins}
\usepackage[binary-units=true]{siunitx}
\DeclareSIUnit{\nothing}{\relax}
\DeclareSIUnit{\ops}{\,operations}
\DeclareSIUnit{\parameters}{\,parameters}
\usepackage{textgreek}
\usepackage[originalparameters]{ragged2e}  
\usepackage[super]{nth}
\definecolor{pastelyellow}{rgb}{0.99, 0.99, 0.59}
\usepackage[obeyFinal,backgroundcolor=pastelyellow,linecolor=black,textwidth=2in]{todonotes}
\reversemarginpar
\ifdraft
\makeatletter
\renewcommand{\todo}[2][]{\@todo[caption={#2}, #1]{\begin{spacing}{1.0}#2\end{spacing}}}
\makeatother
\fi

\newcolumntype{r}[1]{>{\RaggedRight}p{#1}}

\allowdisplaybreaks

\newcommand{\insertfig}[3]{%
   \begin{figure}[!tp]
      \centering
      \includegraphics[width=\textwidth]{#2}
      \caption{#3}
      \label{#1}
   \end{figure}}

\newcommand{\insertfigd}[3]{%
   \begin{figure*}[!tp]
      \begin{adjustwidth}{-2.25in}{0in}
      \begin{minipage}{\textwidth+2.25in}
         \centering
         \includegraphics[width=\textwidth]{#2}
         \caption{#3}
         \label{#1}
      \end{minipage}
      \end{adjustwidth}
   \end{figure*}}

\newcommand{\insertsubtables}[3]{%
   \begin{table}[!tp]
      \begin{minipage}{\columnwidth}
         \caption{#2}
         \label{#1}
         \centering
         \renewcommand{\arraystretch}{1.15}
         #3
      \end{minipage}
   \end{table}}

\newcommand{\inserttab}[4]{%
   \begin{table}[!tp]
      \begin{minipage}{\columnwidth}
         \caption{#3}
         \label{#1}
         \centering
         \renewcommand{\arraystretch}{1.15}
         \begin{tabular}{#2}
            \hline
            #4
            \hline
         \end{tabular}
      \end{minipage}
   \end{table}}

\newcommand{\inserttabd}[4]{%
   \begin{table*}[!tp]
      \begin{adjustwidth}{-2.25in}{0in}
      \begin{minipage}{\textwidth+2.25in}
         \caption{#3}
         \label{#1}
         \centering
         \renewcommand{\arraystretch}{1.15}
         \begin{tabular}{#2}
            \hline
            #4
            \hline
         \end{tabular}
      \end{minipage}
      \end{adjustwidth}
   \end{table*}}

\newcommand{\insertalg}[3]{%
   \begin{algorithm}[!tp]
      \caption{#2}
      \label{#1}
      \DontPrintSemicolon
      #3
   \end{algorithm}}

\begin{document}


\vspace*{0.2in}

\begin{flushleft}
{\Large
\textbf\newline{Automated segmentation of microtomography imaging of Egyptian mummies}
}
\newline
\\
Marc~Tanti\textsuperscript{1\textcurrency},
Camille~Berruyer\textsuperscript{2},
Paul~Tafforeau\textsuperscript{2},
Adrian~Muscat\textsuperscript{1},
Reuben~Farrugia\textsuperscript{1},
Kenneth~Scerri\textsuperscript{3},
Gianluca~Valentino\textsuperscript{1},
V.~Armando~Solé\textsuperscript{2},
Johann~A.~Briffa\textsuperscript{1*}
\\
\bigskip
\textbf{1}
      Dept.\ of Comm.\ \& Computer Engineering,
      University of Malta, Msida MSD 2080, Malta
\\
\textbf{2}
      European Synchrotron Radiation Facility,
      71 Avenue des Martyrs, CS-40220, 38043 Grenoble cedex 09, France
\\
\textbf{3}
      Dept.\ of Systems \& Control Engineering,
      University of Malta, Msida MSD 2080, Malta
\\
\bigskip

%
%


\textcurrency Current Address:
      Institute of Linguistics \& Language Technology,
      University of Malta, Msida MSD 2080, Malta



* johann.briffa@um.edu.mt

\end{flushleft}


\section{Abstract}

Propagation Phase Contrast Synchrotron Microtomography (PPC-SRμCT) is
the gold standard for non-invasive and non-destructive access to internal
structures of archaeological remains.
In this analysis, the virtual specimen needs to be segmented to separate
different parts or materials, a process that normally requires considerable
human effort.
In the Automated SEgmentation of Microtomography Imaging (ASEMI) project,
we developed a tool to automatically segment these volumetric images, using
manually segmented samples to tune and train a machine learning model.
For a set of four specimens of ancient Egyptian animal mummies we achieve an
overall accuracy of 94--98\% when compared with manually segmented slices,
approaching the results of off-the-shelf commercial software using deep
learning (97--99\%) at much lower complexity.
A qualitative analysis of the segmented output shows that our results are
close in terms of usability to those from deep learning, justifying the use
of these techniques.

\ifdraft
\linenumbers
\fi



\section{Introduction}


For at least a decade, archaeologists have been interested in applying new
technologies to their research to better understand our past.
State-of-the-art Propagation Phase Contrast Synchrotron Microtomography
(PPC-SR\textmugreek{}CT) is becoming a golden standard for a non-invasive
and non-destructive access to internal structures of archaeological remains.
This technique has been recently applied to archaeozoological studies of
mummified animal remains from the Ptolemaic and Roman periods of ancient Egypt
(around \nth{3} century BC to \nth{4} century AD).
Researchers performed virtual autopsies and virtual unwrapping, uncovering
information about animal life and death in past civilisations, as well
as revealing the processes used to make these mummies \cite{Berruyer2020,
Porcier2019}.
However, this can be a long process.
After microtomographic data processing and reconstruction, the virtual
specimen has to be segmented to separate the different parts or different
materials of the sample.
For biological samples, such as animal mummies, segmentation is usually done
semi-manually, and can require weeks of human effort, even for small volumes.
Effective automatic segmentation, most likely based on Artificial Intelligence
(AI), could drastically reduce this effort, with the expected computation
time reduced to a matter of hours, depending on the size and complexity of
the volumetric image.
This is particularly relevant as the number and sizes of volumes increases.
For example, the European Synchrotron Radiation Facility (ESRF) is currently
generating huge amounts of data of human organs (up to 2\,TiB for a single
scan); only AI approaches can handle segmentation at this scale.


In computer vision, image segmentation is an important step in providing
the input to higher level tasks such as image understanding, by dividing an
image into meaningful regions.
Automatic segmentation algorithms have been developed for CT images in the
domains of medical imaging, geology and materials analysis, with recent
works considering the use of both supervised and unsupervised techniques in
3D tomography.
Classical methods, using unsupervised techniques, are generally based on
clustering \cite{vrooman2006} or deformable models \cite{cremers2007}, with
the latter being more resilient to inter-slice deformations and intrinsically
accommodate 3D models.
In an example from materials analysis \cite{zhang2017}, cement microtomography
images were segmented using self-organizing maps based on neighbourhood
features, outperforming K-means clustering and the classical edge operators.
Geometric segmentation in 2D and 3D was used in \cite{loeffler2018}
to exploit differences in the geometry of crack formation as opposed to
naturally occurring voids.
In medical imaging, pelvic bone segmentation from 2D CT slices was reported in
\cite{chandar2016}, where segmentation was achieved following a series of steps
which included enhancement, median filtering, smoothing, and contour detection.
A review of techniques used for segmentation of the pelvic cavity is available
in \cite{ma2010}.
Clustering was also used in \cite{majd2010} for segmentation of CT images
containing abdominal aortic aneurysms, using spatial fuzzy C-means.

More recently, fully convolutional networks (FCN) were trained to process
3D images in order to produce automatic volumetric semantic segmentation of
medical images \cite{haberl2018, Ronneberger2015}.
The area of biomedical image segmentation is of particular interest to
this project because in both cases there is limited availability of ground
truth data.
This poses a significant challenge for the application of deep learning
techniques, which generally require a considerable amount of training data.
In an attempt to counteract this problem, data augmentation with elastic
deformations was used in \cite{Ronneberger2015} to augment the training
data set.
Unfortunately, there is a significant difference between conventional CT
scans and the volumetric images captured by microtomography, both in terms
of resolution and contrast.
This means that algorithms developed for conventional CT scans are unlikely
to work in our application without significant adaptation.
Finally, while research in semantic segmentation of 2D images and videos
is mature \cite{badrinarayanan2017}, methods applicable to 3D imagery are
limited \cite{qi2017, wang2018}.
These methods are not directly applicable to our application, as they consider
only 3D point cloud data rather than volumetric scans (voxels).
Furthermore, these algorithms were trained and optimized to segment indoor
scenes or street scenes, rather than CT images.

Automatic segmentation of volumetric images is most well-developed in medical
imaging, such as the CHAOS Challenge \cite{Kavur2020} task for segmenting
the liver from a human CT scan, where many of the top-ranking methods are
variations on the deep learning neural network U-Net \cite{Ronneberger2015}.
By contrast, in archaeology, the problem is less well-studied, and the
techniques available are less sophisticated.
For example, Hati et al.\ \cite{Hati2020} segmented an entire human mummy
into jewels, body, bandages, and the wooden support frame, using a combination
of voxel intensity and the manual selection of geodesic shapes.
This method has the disadvantage that it is only suitable for segmenting
elements that have a high contrast or well-defined edges.
For example, the method in \cite{Hati2020} did not distinguish between soft
tissue, bones, and teeth within the mummy.
Low resolution volumetric scans of mummies were considered in
\cite{OMahoney2019, Friedman2012}, with segmentation performed one slice at
a time using classical machine learning techniques.
In both works, ridged artifacts and loss of detail were evident in the
results obtained.

The automatic segmentation in commercially available software for volumetric
editing, such as Dragonfly~\cite{dragonfly}, requires a human expert to
manually segment a small subset of slices from the tomography image with
all the parts that need to be identified.
A machine learning model is then trained from the manually segmented examples
and used to segment the rest of the volume.


In the Automated SEgmentation of Microtomography Imaging (ASEMI) project,
we developed a fully automatic segmenter using the same methodology, based
on classical machine learning, which we present here.
Our approach has a number of advantages over the state of the art, both in
scientific publications and commercial solutions.
The main contributions of our work are the following:
\begin{itemize}
\item Our segmenter is not limited by the computer's main memory, allowing
it to segment arbitrarily large volumes
(e.g.\ the largest scan presented here requires \SI{133}{\gibi\byte} for
the volume itself, and a further \SI{66}{\gibi\byte} for the labels, while
the feature vectors would require several \si{\tebi\byte}).
\item The parameters of our segmenter are automatically optimised for the
given volume, minimising user input.
\item Our segmenter works directly with three dimensional features, rather
than reducing the problem to an independent segmentation of two-dimensional
slices, without increasing computational complexity from quadratic to cubic.
This allows our system to scale well, particularly for larger volumes.
\item Our system can use interpretable machine learning models, making it
possible to inspect the reasons behind segmentation errors.
\end{itemize}


Experimental results based on four specimens of animal mummies have an
overall accuracy of 94--98\% for our proposed system, when compared with
manually segmented slices.
This approaches the results of off-the-shelf commercial software using
deep learning (97--99\%) at much lower complexity.
A qualitative analysis of the outputs shows that our results are close in
terms of usability to those from deep learning.
Some postprocessing is necessary to clean up segmentation boundaries, for
both our system and the deep learning approach.


Advances in automatic segmentation can be directly applied to a wide variety
of other application domains where microtomography is an important tool.
In industrial applications automatic segmentation will be useful in
non-destructive metrology and detection of voids, cracks and defects.
In biomedical research, it would be useful in small animal and tumour imaging,
amongst others.
Other applications include nanotechnology, geology, electronics and food
science.
The​ main objective of this work is to develop and use artificial
intelligence techniques to automatically segment volumetric microtomography
images of Egyptian mummies, and label each voxel as textiles, organic tissues
or bones.
The main advantages of our proposed system are that it can achieve high
accuracy at lower complexity, allowing it to scale well for very large volumes.


All software and datasets used to generate the results in this work are
available for free download.
The input and output datasets are available under a Creative Commons
Attribution-NonCommercial-ShareAlike 4.0 International License at the
ESRF heritage database for palaeontology, evolutionary biology and
archaeology~\cite{asemi-datasets}.
The source code has been released under the GNU General Public Licence (GPL)
version 3 (or later) and can be found, together with its documentation,
on GitHub~\cite{asemi-segmenter}.




\section{Algorithm design}

\subsection{Overview}

The ASEMI segmenter uses an algorithm that determines, for every voxel in a
volumetric image, a label that corresponds to the material or part that the
voxel belongs to.
For any given volumetric image there may be a number of labels $N \geq 2$,
where one of the labels represents those parts or materials that are not
further distinguished (this label can be interpreted as `background' or
`anything else').
The algorithm works by first obtaining a descriptive feature vector for
every voxel, and then using a classifier to determine a label based on the
feature vector.

\subsection{The voxel neighbourhood}

The feature vector for a given voxel depends on the voxel itself and also
on the voxel's context, which is necessary to determine the texture that a
voxel forms part of.
Formally, the feature vector is a function of the voxel neighbourhood at a
given scale, which is defined as follows.
The neighbourhood of radius $r$ is the cube of voxels of side $2r+1$ centered
around the voxel of interest.
The scale $s$ is the number of times the entire volume is resized by half, so
that a neighbourhood can be from the original volume ($s=0$) or from a smaller
version of the volume at the corresponding location of the central voxel.
These relationships are illustrated in Fig~\ref{fig:neighbourhood-scales}.
\insertfig{fig:neighbourhood-scales}
    {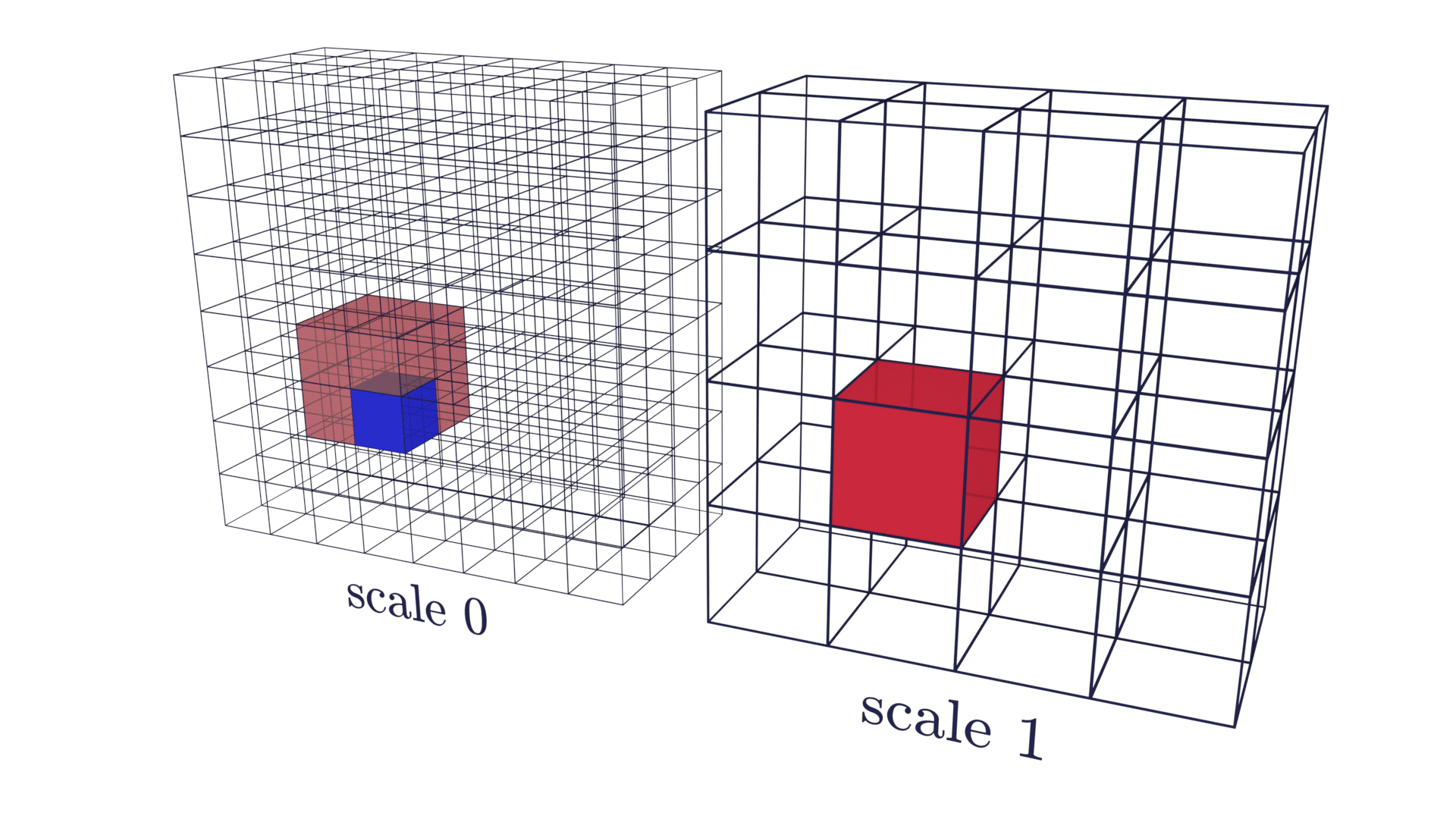}
    {{\bf Voxels at different scales.}
    An illustration of a voxel at scale 0 (blue) and its corresponding voxel
    at scale 1 (red).
    A voxel at scale 1 (red) corresponds to every voxel in a $2\times 2\times
    2$ region at scale 0 (light red).}
The scaled volumes are obtained by first applying a 3D Gaussian blur to the
entire volume and then decimating this by a factor of $2^s$ in each dimension.
This creates versions of the volume with less detail, allowing the algorithm
to operate on larger-scale textures.

\subsection{Feature vectors}

For any given voxel neighbourhood radius and scale, a number of different
features may be obtained.
We usually use a concatenation of several feature vectors, each obtained at
a different neighbourhood radius and scale, as an input to the classifier.

The simplest feature is the intensity value of the voxel being classified,
which corresponds to the density of the material in that location.
Clearly, this feature is independent of the neighbourhood radius, and does not
carry any information about the material texture.
While it is possible to consider the intensity at different scale, we usually
use this feature only at scale zero.
Therefore, no parameters are necessary to uniquely define this feature.

Next up in complexity is the histogram of intensities within the given
neighbourhood.
This feature vector consists of an array of $k$ frequencies, one for each
linearly-spaced bin of voxel intensities.
This contains information about the distribution of material density in the
given neighbourhood.
Computation of this feature vector is defined by three parameters: the
neighbourhood radius $r$, scale $s$, and number of bins $k$.

The distribution of material density is one aspect of the texture that a
voxel forms part of.
Another aspect is the structure of the texture.
This is captured well by the Local Binary Pattern (LBP)~\cite{Ojala1996},
which operates on a plane, and is insensitive to density variation but
sensitive to the relative location of voxel values.
In order to capture information in three dimensions, we combine
the features from three orthogonal planes centered on the voxel
under inspection \cite{Zhao2007, Abbasi2017}, as illustrated in
Fig~\ref{fig:neighbourhood-planes} for a neighbourhood of radius $r=2$.
\insertfig{fig:neighbourhood-planes}
    {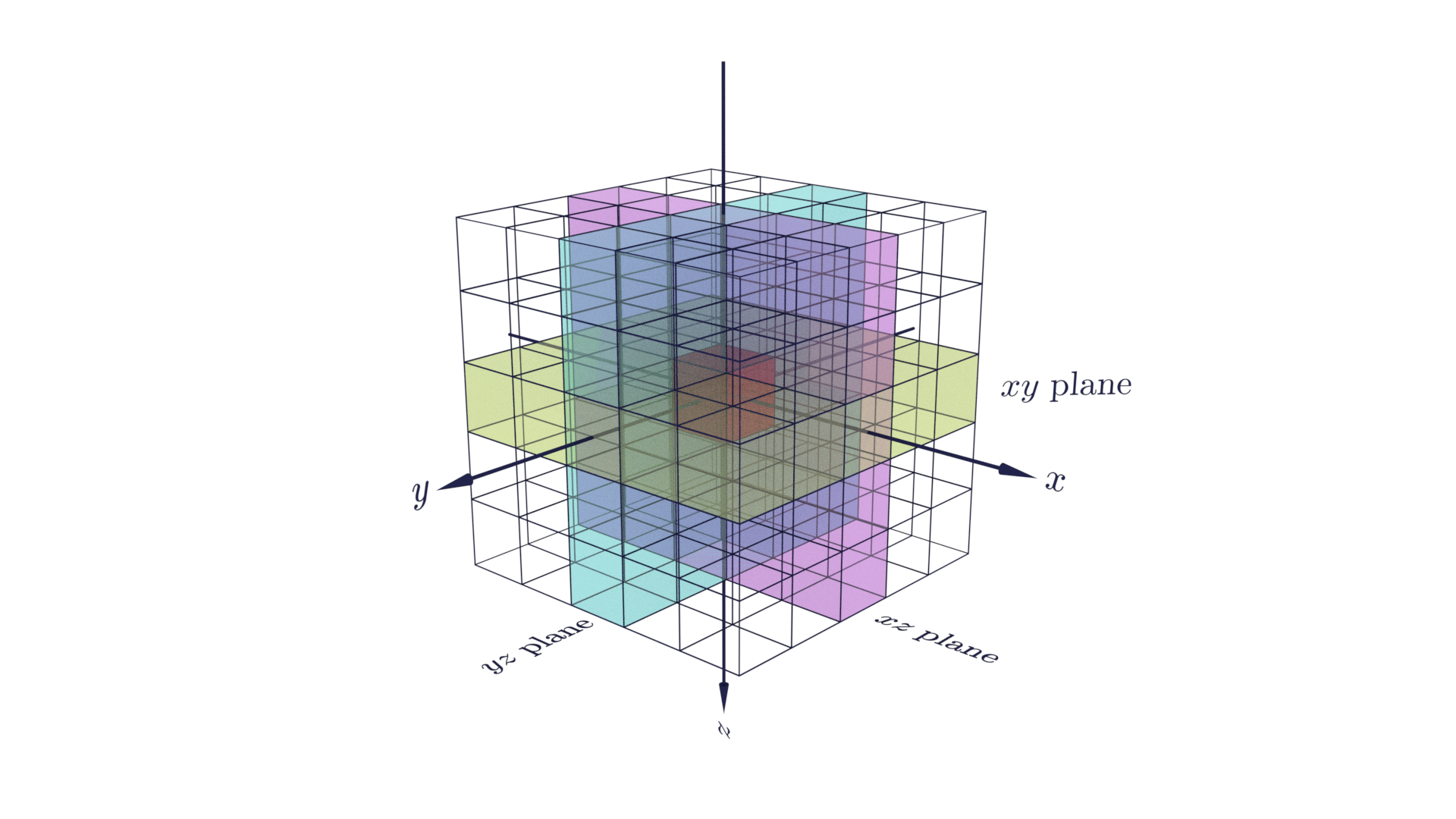}
    {{\bf Three orthogonal planes within a voxel neighbourhood.}
    An illustration of the three orthogonal planes in the neighbourhood of
    radius $r=2$ for the red voxel in the center.
    The $xy$, $yz$, and $xz$ planes are shown in magenta, cyan, and yellow
    respectively.}
First, an LBP code is obtained for every voxel, based on its immediate
$3\times3$ neighbourhood in the plane considered (i.e.\ the 8 neighbours at
radius 1).
Uniform and rotation invariant LBP codes~\cite{barkan2013} are used, as
preliminary experiments showed that these improve performance while reducing
the feature vector size.
Next, we obtain a histogram of the codes corresponding to voxels in a 2D
neighbourhood of radius $r$ in the plane considered, and use this as our
feature vector.
Computation of this feature vector is defined by three parameters: the scale
$s$ and histogram neighbourhood radius $r$.
The number of histogram bins is fixed to 10, as that is the number of
different codes possible.

Other features exist for encoding textural information, such as
SIFT~\cite{lowe1999} and SURF~\cite{bay2008}, which have found extensive
use in computer vision.
In preliminary experiments we found that these features were computationally
much more expensive and did not improve performance.
Therefore, we did not consider them further for this application.

The final feature vector chosen is the concatenation of:
\begin{inparaenum}[a)]
    \item the value of the voxel being classified,
    \item a short range histogram of voxel values at scale zero,
    \item a long range histogram of voxel values (at a scale to be determined), and
    \item a three orthogonal planes LBP (at a scale to be determined).
\end{inparaenum}
Preliminary experiments showed that the combination of histogram and LBP
features was better than either one alone.
We also experimented with two-dimensional histograms across the three
orthogonal planes, as well as the mean value of the voxel neighbourhood, but
the performance increase was minimal and did not justify the added complexity.
Using both short and long range features (both histogram and LBP) resulted in
a small performance increase over using only long range features.
Rather than use long and short range features for both histogram and LBP, we
opted to limit this to the histogram features which are faster to compute.

\subsection{Classifier}

The extracted features are used as input to a classifier, which is trained
to create a mapping between the feature vectors and the corresponding label.
Two classifiers are used in this work: the random forest
classifier~\cite{ho1998}, as implemented in scikit-learn~\cite{scikit-learn},
and the feedforward neural network~\cite{bishop1995} implemented in
Tensorflow~\cite{tensorflow2015-whitepaper}.
The random forest classifier has the advantage of being an interpretable
model while the neural network generalises well and has a faster GPU-based
implementation.
We also experimented with support vector machines~\cite{cristianini2000},
but these were not performing as well as the other methods so we did not
consider them further.

The random forest classifier was trained with a node splitting criterion
based on the Gini impurity, no limit on the number of leaf nodes, a maximum
tree depth of 16, and the number of features per tree equal to the square
root of the total number of features.
The only remaining hyperparameter, the number of trees, was left as a free
variable to be determined in the hyperparameter optimisation process, within
the range specified in Table~\ref{tbl:design_optimisation}.
\insertsubtables{tbl:design_optimisation}
    {\bf Search space for free variables in feature selection and model
    hyperparameters.}
    {
    \begin{subtable}[b]{\linewidth}
        \centering
        \caption{Feature selection.}

        \begin{tabular}{ll|c}
            \hline
            \emph{Feature} & \emph{Variable} & \emph{Search Space} \\
            \hline
            Histogram 1 & Radius & 1-8 \\
            & Bins & \{8, 16, 32\} \\
            \hline
            Histogram 2 & Scale & 0-2 \\
            & Radius & 1-32 \\
            & Bins & \{8, 16, 32\} \\
            \hline
            LBP & Scale & 0-2 \\
            & Radius & 1-32 \\
            \hline
        \end{tabular}
    \end{subtable}
    \par\medskip

    \begin{subtable}[b]{\linewidth}
        \centering
        \caption{Model hyperparameters.}

        \begin{tabular}{ll|c}
            \hline
            \emph{Classifier} & \emph{Hyperparameter} & \emph{Search Space} \\
            \hline
            Random forest & Number of trees & \{16, 32, 64\} \\
            \hline
            Neural network & Layer 1 size & \{32, 64, 128, 256\} \\
            & Layer 2 size & \{32, 64, 128, 256\} \\
            & Dropout rate & 0.0-0.5 \\
            & Initialisation stddev. & 0.0001-1.0 \\
            & Minibatch size & \{4, 8, 16, 32, 64\} \\
            \hline
        \end{tabular}
    \end{subtable}
    }

Similarly, the neural network structure included two hidden layers with a
leaky ReLU activation function, and was trained with the Adam optimiser,
early stopping criterion, dropout, and weight initialisation using normally
distributed random numbers with zero mean and bias initialisation of zero.
The remaining hyperparameters, left as free variables to be
determined in the hyperparameter optimisation process, are listed in
Table~\ref{tbl:design_optimisation}, together with the range explored.

While other more complex hyperparameter optimisation approaches have been
used in the literature, in this work we opted for a simple approach.
Our optimisation process consists of a two-stage random search in the combined
search space of the feature selection variables and model hyperparameters.
The first stage randomly samples the entire search space, uniformly across
all variables; this explores the search space more efficiently than a grid
search at the same resolution.
The second stage is a local search, starting with the best result from the
first stage.
The algorithm randomly changes a single variable, in an iterative process
that is similar to a stochastic hill climbing optimiser.
Details are outlined in Algorithm~\ref{alg:design_search}.
\insertalg{alg:design_search}
    {\bf Two-stage search algorithm.}
    {
    \KwIn{A training set and a validation set.}
    \KwData{A number of global iterations $I_g = 1000$, a number of local
    iterations $I_l = 10000$, a number of training set voxel samples
    $S_t = 16000$, a number of validation set voxel samples $S_v = 16000$, a
    number of seconds as a timeout limit $t = 5$.}
    \KwOut{An optimised set of variables}

    sampledTrainSet $\leftarrow$ a sample of $S_t$ items from each label in the training set

    sampledValSet $\leftarrow$ a sample of $S_v$ items from each label in the validation set

    knownVarSets $\leftarrow$ $\varnothing$

    bestVarSet $\leftarrow$ null

    bestAccuracy $\leftarrow$ 0

    \Begin(global phase){
        \For{$I_g$ times}{
            \Repeat{varSet $\notin$ knownVarSets}{
                varSet $\leftarrow$ a random set of variables

                \lIf{more than $t$ seconds have passed}{exit global phase}
            }

            add varSet to knownVarSet

            segmenterModel $\leftarrow$ new segmenter model using varSet

            train segmenterModel using sampledTrainSet

            modelAccuracy $\leftarrow$ evaluate segmenterModel accuracy using sampledValSet

            \If{modelAccuracy $>$ bestAccuracy}{
                bestVarSet $\leftarrow$ varSet

                bestAccuracy $\leftarrow$ modelAccuracy
            }
        }
    }

    \Begin(local phase){
        \For{$I_l$ times}{
            \Repeat{varSet $\notin$ knownVarSet}{
                varSet $\leftarrow$ bestVarSet

                replace a single variable in varSet with a random value

                \lIf{more than $t$ seconds have passed}{exit local phase}
            }

            add varSet to knownVarSet

            segmenterModel $\leftarrow$ new segmenter model using varSet

            train segmenterModel using sampledTrainSet

            modelAccuracy $\leftarrow$ evaluate segmenterModel accuracy using sampledValSet

            \If{modelAccuracy $>$ bestAccuracy}{
                bestVarSet $\leftarrow$ varSet

                bestAccuracy $\leftarrow$ modelAccuracy
            }
        }
    }

    \KwResult{bestVarSet}
    }


\section{Implementation Efficiency and Scalability}

\subsection{Overview}

For the application considered, with microtomography images of $10^9$ to
$10^{10}$ voxels, the scalability of the algorithms is a critical concern.
There are two aspects to this: memory requirement and computational complexity.
At this scale, the images are at the limit of what will fit in current
high-RAM machines, while the set of feature vectors for the entire volume
will easily exceed available memory.
Next-generation scans, such as those recently started at the ESRF, generate
images that are already larger than available memory.
It is therefore necessary for the algorithm implementation to divide the
volume into subvolumes and operate on these separately.
In a naive implementation, the computational complexity scales linearly with
the number of voxels, as the extraction of the feature vector can be seen
as independent for each voxel.
However, neighbouring voxels have overlapping neighbourhoods, so that some
computations can be shared between them.
An optimised implementation can leverage this to reduce the overall complexity.

\subsection{Optimised histogram computation}


The main contributor to the computational complexity of feature extracation
is the computation of histograms of voxel neighbourhoods, used for both the
histogram feature vector and the LBP feature vector.
When the histogram for each neighbourhood is independently computed, the
complexity is cubic with respect to the neighbourhood radius $r$.
If we incrementally compute histograms for the neighbourhoods of adjacent
voxels, however, we can reduce the overall complexity to quasi-quadratic.

Consider the neighbourhood of radius $r$ of a voxel $v_{x,y,z}$ at index
$x,y,z$ in the volume.
Computing the histogram for this single neighbourhood requires consideration of
$(2r+1)^3$ voxel intensities, for a complexity $O(r^3)$.
Now consider the neighbourhood of adjacent voxel $v_{x,y,z+1}$.
This overlaps with the neighbourhood of $v_{x,y,z}$, except for the planes
$z-r$ and $z+r+1$, as illustrated in Fig~\ref{fig:neighbourhood-overlap}.
\insertfig{fig:neighbourhood-overlap}
    {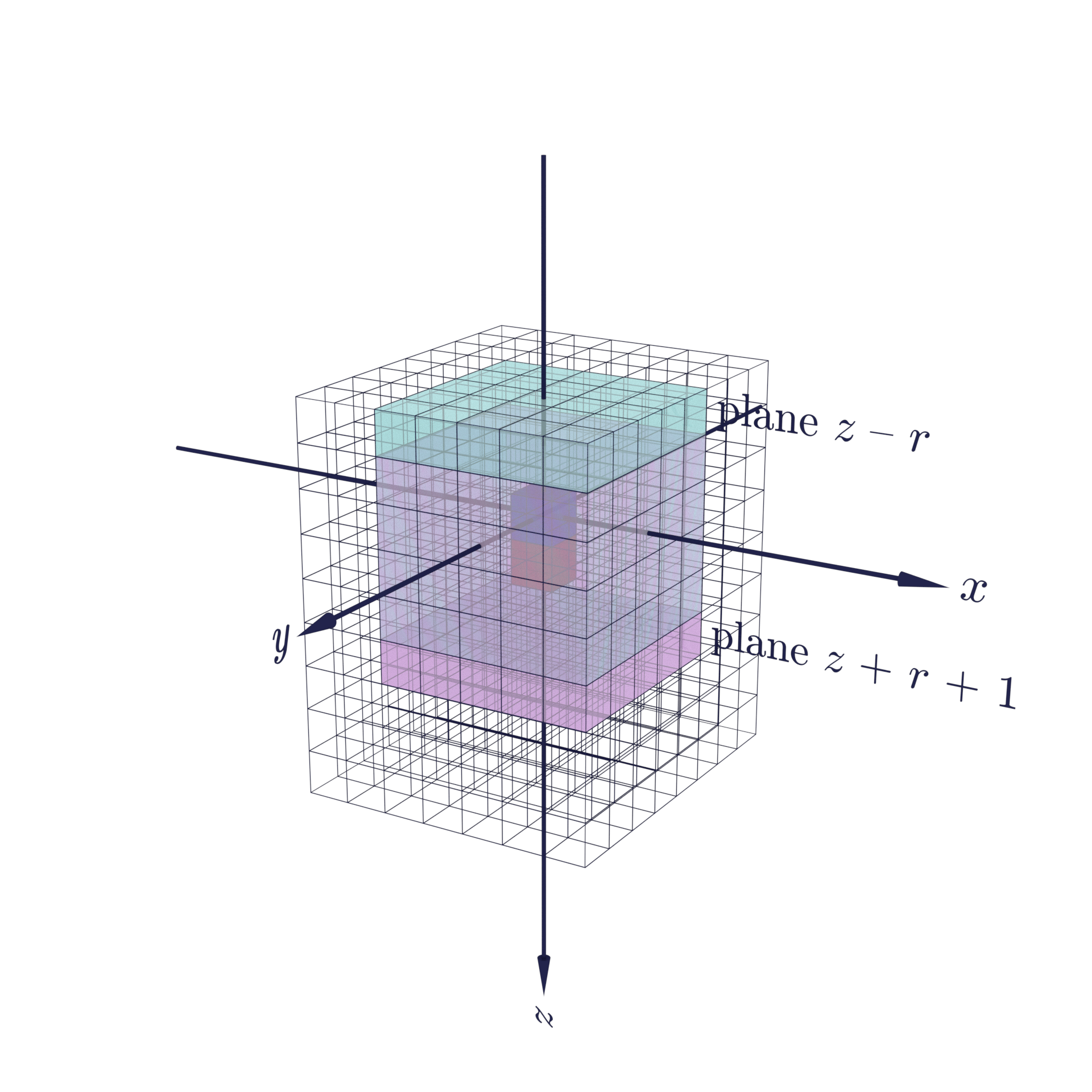}
    {{\bf Neighbourhood overlap for adjacent voxels.}
    An illustration of the overlap between the neighbourhoods of radius $r=2$
    for the blue voxel $v_{x,y,z}$ (shown in cyan), and
    the adjacent red voxel $v_{x,y,z+1}$ (shown in magenta).
    The plane $z-r$ exists only in the first neighbourhood, while
    the plane $z+r+1$ exists only in the second neighbourhood.}
Therefore, we can compute the neighbourhood histogram for $v_{x,y,z+1}$ as an
incremental change on that of $v_{x,y,z}$.
Starting with the histogram for $v_{x,y,z}$, we subtract the frequencies
for the plane $z-r$ and add the frequencies for the plane $z+r+1$.
Each of these two operations requires consideration of $(2r+1)^2$ voxel
intensities, for an overall complexity $O(r^2)$ for this incremental histogram
computation.

For any given subvolume, the first neighbourhood histogram needs to be computed
independently, while all other histograms can be determined incrementally.
Assuming the subvolume size is much larger than $r$, the effect on complexity
of the first histogram becomes insignificant, for an overall complexity
approaching $O(r^2)$.


A considerable improvement in efficiency is also obtained by computing
histograms on a GPU, using NVIDIA's CUDA framework \cite{cuda-pg-10-0}.
For incremental histogram computation it is advantageous to consider operations
on independent slices.
We can compute the histograms for each voxel neighbourhood in parallel, because
they are independent.
This allow us to take advantage of the massively parallel GPU architecture.
Furthermore, observe that even in this case, the neighbourhoods for adjacent
voxels overlap significantly.
We take advantage of this overlap by loading the neighbourhood into on-chip
`shared' memory, which is accessible at cache-level speeds by the computational
units of the multiprocessor.
This significantly reduces the number of memory reads required, for a
significant speedup in this I/O-bound problem.

We also take advantage of a number of GPU architecture features to ensure
an efficient implementation.
These include the parallel reading of tile neighbourhood data into shared
memory, organised so that access by warp threads is on different memory banks.
Incrementally computed histograms are also kept in shared memory, writing only
the final 3D histogram results to main memory.


Note that these improvements are also of benefit when computing LBP features.
Specifically, for each of the three orthogonal planes, once the LBP codes
corresponding to each voxel are computed, a histogram of these codes is
obtained for the 2D voxel neighbourhood of radius $r$.
The computation of the LBP codes takes constant time, as for each voxel it
requires a comparison with the eight immediate neighbours.
Computing the histogram of the codes requires consideration of $(2r+1)^2$
voxel intensities, for a complexity $O(r^2)$.
The histogram computation therefore dominates the complexity of the LBP
feature extraction.


Comparative timings were obtained during development to validate these
optimisation steps.
Table~\ref{tbl:speedups} gives the time taken for the feature extraction
process on a small $180\times150\times200$ volume using an Intel Core i7-8700K
CPU at \SI{3.7}{\giga\hertz} and an NVIDIA GeForce GTX\,1080\,Ti.
\inserttab{tbl:speedups}{l|cc}
    {{\bf Validating speedups on a small volume.}
    Volume size $180\times150\times200$;
    CPU: Intel Core i7-8700K at \SI{3.7}{\giga\hertz};
    GPU: NVIDIA GeForce GTX\,1080\,Ti.}
    {
    & CPU (\si{\second}) & GPU (\si{\second}) \\
    \hline
    Histogram ($r=32, s=0, k=32$)   & 17.8 & 1.1 \\
    LBP (3-plane, $r=32, s=0$)      & 24.4 & 1.3 \\
    }
In all cases, histograms are computed using the incremental algorithm.

\subsection{Complexity analysis and comparison with U-Net}

The overall complexity of the ASEMI segmenter can be determined by considering
both the feature extraction and the classifier.
The complexity for computing the features depends directly on the neighbourhood
radius $r$, as already shown.
When incremental histogram computation is used, the complexity for both the
3D histogram and the three-orthogonal-plane LBP features have a complexity
$O(r^2)$.
The complexity of the classifier depends on the number of features computed,
as well as on the statistical properties of the data (i.e.\ how easily
separable the data is).
The number of features is equal to the number of bins $k$ for the histogram
features, and has a fixed value of 30 for the three-orthogonal-plane LBP.
Due to the dependency on the statistical properties of the data, we determine
the complexity for the classifiers empirically, for the cases considered in
the results section.
Typical models based on the ASEMI segmenter (random forest and three layer
neural network) require around \SIrange{20}{60}{\kilo\ops} per voxel,
including the feature computation.

In the U-Net architecture, the size of the context over which the
classification decision is computed is taken as equal to the effective
receptive field.
Since the effective receptive field depends on the complete architecture of
the network we compare the complexity empirically for typical architectures
from the literature.
The 3D U-Net \cite{Cicek2016} is characterised with
\SI{19.07}{\mega\parameters} and a receptive field of approximately 88 voxels
in each direction, equivalent to $r \approx 43$.
Considering only multiplication operations, we determined the number of
computations per output voxel to be around \SI{5.4}{\mega\ops} per voxel.
For the 2D U-Net \cite{Ronneberger2015}, around \SI{640}{\kilo\ops} per
voxel are required when $r \approx 30$.

It is clear from this comparison that the ASEMI segmenter has a significant
complexity advantage over deep learning methods.
In both cases, much of the computation can be performed on a GPU accelerator.
Furthermore, the feature vector sizes of a typical deep neural network are
in the order of hundreds whereas our feature vector sizes are in the order
of tens, giving our algorithm a potential advantage in GPU memory requirements.
These results justify the use of classical methods over deep learning methods
for 3D segmentation.


\section{Experimental setup}

\subsection{Overview}

Experimental results, based on four specimens, compare the performance of
the ASEMI segmenter with U-Net.
The overall process, from scanning each specimen to its automatic labelling,
is shown in Fig~\ref{fig:system}.
\insertfigd{fig:system}
    {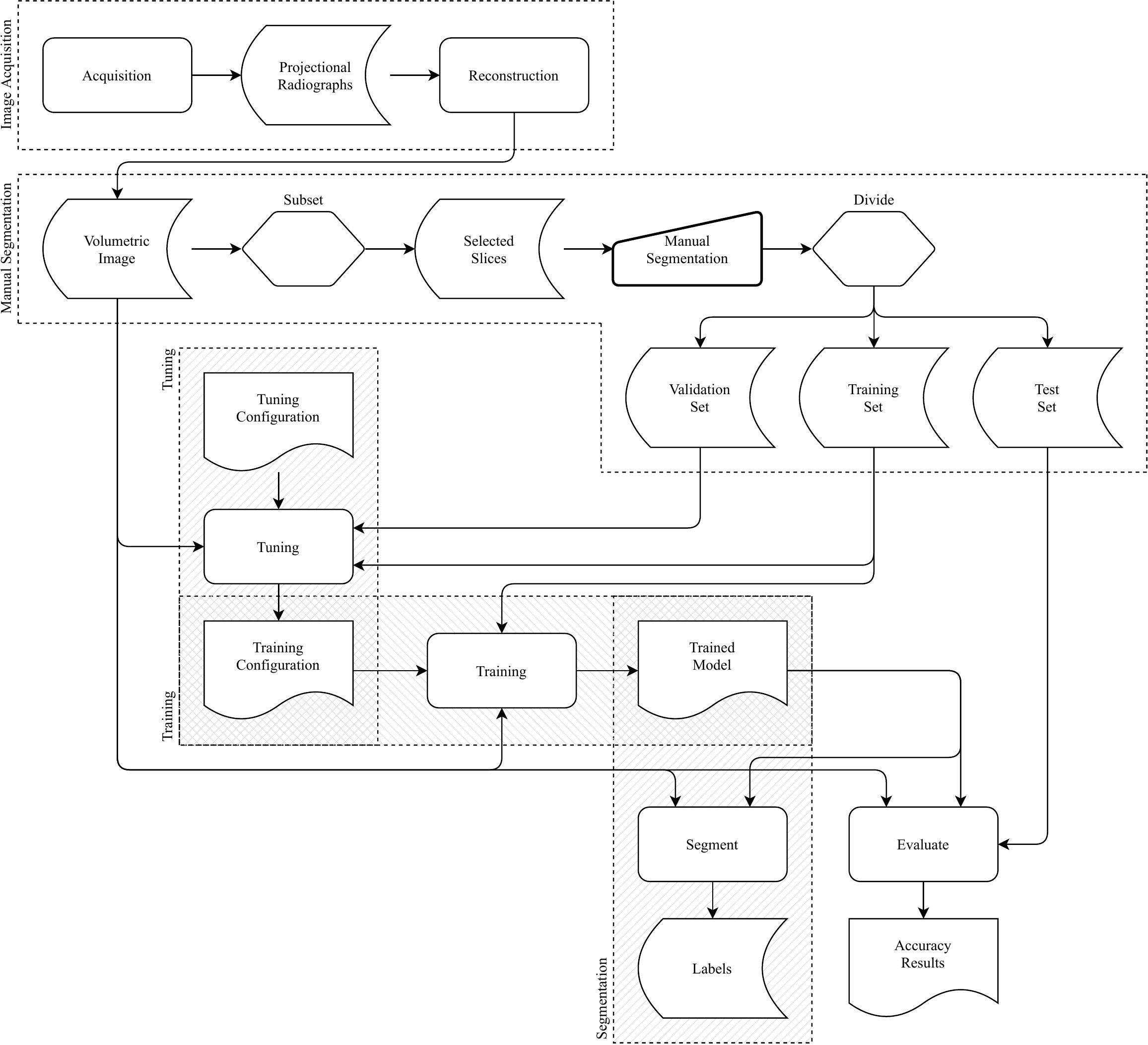}
    {{\bf Block diagram of the overall process, from scanning to automatic
    labelling.}
    }
Following image acquisition and reconstruction, a selection of slices
from each volumetric image are manually segmented and divided into three
independent sets, for training, validation, and testing of the machine
learning models obtained.
For the ASEMI segmenter, a feature selection and model hyperparameter
optimisation process uses a subset of the training and validation sets to
determine the best set of features and classifier parameters, which are then
used to train the machine learning model.
For U-Net, these two steps are combined.
In both cases, the trained model is then used to segment the complete volume.
For final presentation, the automatic segmentations were cleaned and rendered
using VGStudioMax, which is excellent for visualisation and rendering.
The trained model is also evaluated by determining the classification accuracy
in comparison with the previously withheld manually segmented test set.
Details of the specimens used and of each of these steps are given in the
rest of this section.

\subsection{Specimens}

The four specimens used in this study are mummified animals from ancient Egypt.
They are curated by the Museum d’Histoire Naturelle de Grenoble
(Fig~\ref{fig:experiments_specimens}, A, B and C) and the Musée de Grenoble
(Fig~\ref{fig:experiments_specimens}, D).
\insertfigd{fig:experiments_specimens}
    {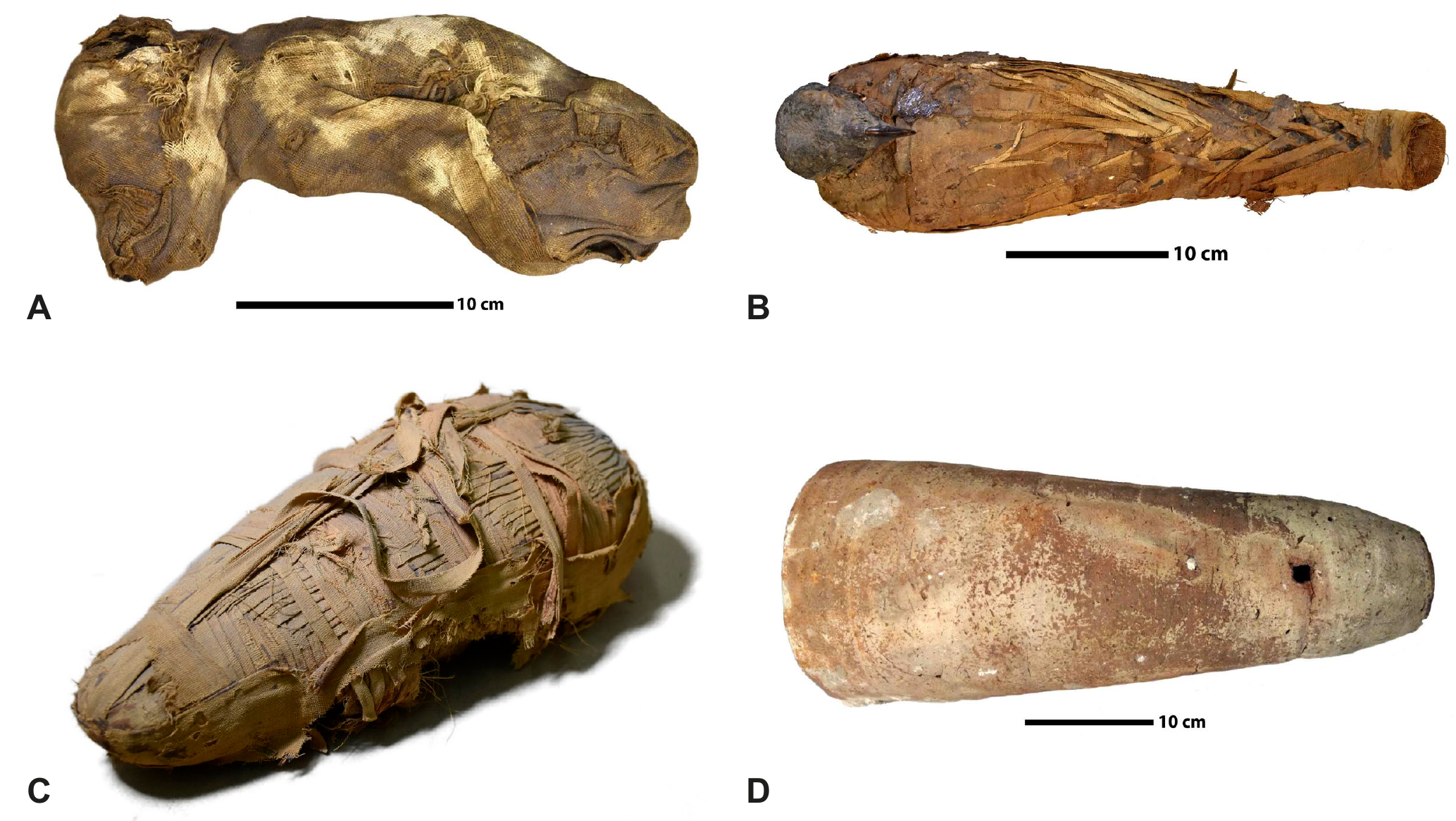}
    {{\bf The four specimens used in our experiments, identified by their
    museum accession numbers.}
    A:~MHNGr.ET.1023 (dog),
    B:~MHNGr.ET.1017 (raptor),
    C:~MHNGr.ET.1456 (ibis),
    D:~MG.2038 (ibis in a jar).}
Like many other animal mummies kept in museums, we know very few things
about their precise historical and archaeological origins.
They are estimated from Ptolemaic and Roman period (around \nth{3} century
BC to \nth{4} century AD).
The first specimen, MHNGr.ET.1023, is a mummy of a puppy.
Traces on its fabrics show that it was wrapped with more bandages than its
actual state.
The second specimen, MHNGr.ET.1456, is the mummy of an ibis still wrapped
in elaborate wrappings.
Third specimen, MG.2038, is the mummy of an ibis enclosed inside a ceramic
jar which was sealed with a type of mortar.
The fourth mummy, MHNGr.ET.1017, is a mummy with a raptor bird's head directly
visible and a wrapped body.
Unlike the other specimens, the volumetric image doesn't contain the whole
mummy, but focuses only on the stomach area of the bird inside the mummy.

\subsection{Image acquisition}

The animal mummies used in this work were scanned at the European Synchrotron
Radiation Facility (ESRF) in Grenoble, France.
The properties of the synchrotron source allows Propagation Phase Contrast
tomography, which gives images with higher resolution and better contrast
than conventional sources.
In our case, three configurations were used, from \SI{24.19}{\micro\meter}
to \SI{50.72}{\micro\meter} voxel size.
For each scan, the volumetric images were reconstructed using a single
distance phase-retrieval algorithm coupled with filtered back projection
implemented in the PyHST2 software package \cite{Mirone2014, Paganin2002}
and ring artefact corrections were applied.
Then, the scanned sections were vertically concatenated and the final files
were saved in 16-bit JPEG2000 format, using lossy compression with a target
compression rate of 10, four levels of wavelet decomposition, and a tile
size that covers the entire image.
Detailed scan parameters are summarized in
Table~\ref{tbl:experiments_scan-details}.
\inserttabd{tbl:experiments_scan-details}
    {r{1.2in}|r{1.8in}r{1.8in}r{1.8in}}
    {\bf Scan parameters used to scan the animal mummies at the ESRF.}
    {
    Voxel size (\si{\micro\meter}) & 50.72 & 47.8 & 24.19 \\
    \hline
    Sample      & MG.2038 (ibis in a jar) complete mummy,
                  MHNGr.ET.1456 (ibis) complete mummy
                & MHNGr.ET.1017 (raptor) complete mummy
                & MHNGr.ET.1023 (dog) complete mummy \\
    Optic       & Lafip 2 – Hasselblad 100
                & 47 micron ID17
                & Lafip 2 \\
    Date        & 12 December 2017
                & 18 November 2017
                & 19 November 2017 \\
    Average detected energy (\si{\kilo\electronvolt})
                & $\approx 129$
                & $\approx 107$
                & $\approx 110$ \\
    Filters (\si{\milli\meter})
                & Cu 6, Mo 0.25
                & Al prof $18 \times 5$, Mo 0.2
                & Al pro  $18 \times 5$, Mo 0.25 \\
    Propagation distance (\si{\milli\meter})
                & \num{4500}
                & \num{4000}
                & \num{3000} \\
    Sensor      & FReLoN 2k14
                & FReLoN
                & PCO Edge 4.2 CLHS \\
    Scintillator
                & Scintillating fiber
                & LuAg 2000
                & LuAg 2000 \\
    Projection number
                & \num{5000}
                & \num{5000}
                & \num{5000} \\
    Scan geometry
                & Half Acquisition 950 pixels offset,
                  vertical scan series with \SI{2.5}{\milli\meter} steps
                & Half Acquisition 500 pixels offset,
                  vertical scan series with \SI{2}{\milli\meter} steps
                & Half Acquisition 300 pixels offset,
                  vertical scan series with \SI{2.3}{\milli\meter} steps \\
    Exposure time (\si{\second})
                & \num{0.04}
                & \num{0.03}
                & \num{0.015} \\
    Number of scan
                & \num{180}
                & \num{191}
                & $126+126$ \\
    Reconstruction mode
                & Single distance phase retrieval \cite{Paganin2002},
                  vertical concatenation, ring artefacts correction,
                  16 bit conversion in JPEG~2000 format, binning
                & Single distance phase retrieval \cite{Paganin2002},
                  vertical concatenation, ring artefacts correction,
                  16 bit conversion in JPEG~2000 format, binning
                & Single distance phase retrieval \cite{Paganin2002},
                  vertical concatenation, ring artefacts correction,
                  16 bit conversion in JPEG~2000 format, binning \\
    }

\subsection{Manual segmentation}

As mentioned earlier, manual segmentation is the most human time-consuming step.
The speed and accuracy of the manual segmentation process depends primarily on
the complexity of the specimen and the quality of its tomographic data, as well
as on the individual performing the segmentation.
In our case, a manual selection with active grey level range was used in order
to segment a selection of slices in Dragonfly.
This step should not be affected by mislabelling or ambiguities except from
human error.

The manually segmented slices were split into a training set, a validation
set, and a test set using 60\%, 20\%, and 20\% of the slices respectively.
The training and validation sets were used to optimise (for ASEMI) and train
the machine learning model, while the test set was used only to determine
the model accuracy.

\subsection{Automatic segmentation with ASEMI}

The process for automatic segmentation with ASEMI starts by optimising the
feature selection and model hyperparameters for each specimen, using a sample
taken from the training and validation sets.
A new model is trained using the optimised features and model hyperparameters
on the full training set.
This model is then evaluated on the full test set, and is also used to
segment the whole volume.

The optimised features and model hyperparameters for each classifier and
specimen are given in Table~\ref{tbl:experiments_optimised_variables}.
\insertsubtables{tbl:experiments_optimised_variables}
    {\bf Optimised features and model hyperparameters.}
    {
    \begin{subtable}[b]{\linewidth}
        \centering
        \caption{Random forest classifier.}
        \begin{tabular}{l l | c c c c}
            \hline
            \emph{Feature} & \emph{Variable}
                & \rotatebox{90}{\emph{\parbox{3cm}{MHNGr.ET.1023\\(dog)}}}
                & \rotatebox{90}{\emph{\parbox{3cm}{MHNGr.ET.1017\\(raptor)}}}
                & \rotatebox{90}{\emph{\parbox{3cm}{MHNGr.ET.1456\\(ibis)}}}
                & \rotatebox{90}{\emph{\parbox{3cm}{MG.2038\\(ibis in a jar)}}} \\
            \hline
            Histogram 1 & Radius & 3 & 6 & 5 & 7 \\
            & Bins & 32 & 8 & 8 & 32 \\
            \hline
            Histogram 2 & Scale & 0 & 0 & 0 & 2 \\
            & Radius & 12 & 27 & 21 & 24 \\
            & Bins & 16 & 32 & 32 & 16 \\
            \hline
            LBP & Scale & 0 & 1 & 1 & 0 \\
            & Radius & 15 & 27 & 20 & 30 \\
            \hline
            \hline
            Random forest & Number of trees & 64 & 64 & 64 & 64 \\
            \hline
        \end{tabular}
    \end{subtable}
    \par\medskip

    \begin{subtable}[b]{\linewidth}
        \centering
        \caption{Neural network classifier.}
        \begin{tabular}{l l | c c c c}
            \hline
            \emph{Feature} & \emph{Variable}
                & \rotatebox{90}{\emph{\parbox{3cm}{MHNGr.ET.1023\\(dog)}}}
                & \rotatebox{90}{\emph{\parbox{3cm}{MHNGr.ET.1017\\(raptor)}}}
                & \rotatebox{90}{\emph{\parbox{3cm}{MHNGr.ET.1456\\(ibis)}}}
                & \rotatebox{90}{\emph{\parbox{3cm}{MG.2038\\(ibis in a jar)}}} \\
            \hline
            Histogram 1 & Radius & 7 & 8 & 7 & 7 \\
            & Bins & 16 & 32 & 8 & 32 \\
            \hline
            Histogram 2 & Scale & 1 & 2 & 0 & 2 \\
            & Radius & 6 & 32 & 27 & 29 \\
            & Bins & 8 & 32 & 32 & 16 \\
            \hline
            LBP & Scale & 1 & 1 & 1 & 0 \\
            & Radius & 18 & 28 & 27 & 28 \\
            \hline
            \hline
            Neural network
            & Layer 1 size & 256 & 256 & 64 & 256 \\
            & Layer 2 size & 64 & 64 & 128 & 64 \\
            & Dropout rate & 0.4 & 0.45 & 0.25 & 0.25 \\
            & Init. stddev. & 0.0001 & 0.0251 & 0.1585 & 0.0016 \\
            & Minibatch size & 32 & 32 & 32 & 32 \\
            \hline
        \end{tabular}
    \end{subtable}
    }
Observe that the random forest always opted for the maximum allowed number
of trees, and the neural network always opted for a minibatch size just
below the maximum allowed.
The LBP and Histogram~2 features tended to select mid-range to large
neighbourhoods, with both classifiers.

\subsection{Automatic segmentation with U-Net}

The deep learning tool in Dragonfly was used to create a 5-level U-Net model
with the required number of output labels.
The model was trained on the same training and validation sets used with
the ASEMI segmenter, and evaluated on the same test set.
For generating the full volume segmentation, however, a separate model
was trained with the full set of manually segmented slices, reserving 25\%
of patches for validation.

Dragonfly does not provide direct tools for hyperparameter optimisation, so we
used the hyperparameters chosen by the ESRF annotators, as follows.
In all cases, the data was augmented with a flip in each direction and also
with a rotation of up to 180\textdegree.
A patch size of 128 was used, with a stride to input ratio of 0.5, batch size
64, the categorical crossentropy loss function, and Adadelta optimization
algorithm, over 100 epochs.

\subsection{Volumetric image parameters and timings}

For each of the four specimens, Table~\ref{tbl:experiments_durations}
shows the size of the volumetric image with the number of slices chosen
for manual segmentation and the number of labels used.
\inserttab{tbl:experiments_durations}{ll|cccc}
    {\bf Volumetric image parameters and time spent on the various stages of
    segmenting the images.}
    {
    &
    & \rotatebox{90}{\emph{\parbox{3cm}{MHNGr.ET.1023\\(dog)}}}
    & \rotatebox{90}{\emph{\parbox{3cm}{MHNGr.ET.1017\\(raptor)}}}
    & \rotatebox{90}{\emph{\parbox{3cm}{MHNGr.ET.1456\\(ibis)}}}
    & \rotatebox{90}{\emph{\parbox{3cm}{MG.2038\\(ibis in a jar)}}} \\
    \hline
    \multirow{3}{1.8cm}{Volume parameters}
    & Volume size ($\times10^9$ voxels)            & 27.2  & 71.2  & 41.5  & 17.1 \\
    & Number of manually segmented slices          & 18/35 & 22    & 21    & 22   \\
    & Number of labels (excluding null)            & 4     & 7     & 8     & 7    \\
    \hline
    \multirow{1}{1.8cm}{Manual}
    & Segmentation time (\si{\hour})               & 4     &       & 6     & 5    \\
    \hline
    \multirow{2}{1.8cm}{U-Net}
    & Training time (\si{\hour})                   & 15    &       & 38    & 20   \\
    & Segmentation time (\si{\hour})               & 3     &       & 12    & 5    \\
    \hline
    \multirow{3}{1.8cm}{ASEMI}
    & Optimisation time (\si{\hour})               & 29.0  & 45.5  & 52.9  & 33.5 \\
    & Training time (\si{\hour})                   & 6.5   & 10.2  & 13.6  & 6.6  \\
    & Segmentation time (\si{\hour})               & 67.7  & 218.7 & 113.6 & 53.0 \\
    \hline
    \multirow{2}{1.8cm}{Post-processing}
    & Time to transfer to VGStudioMax (\si{\hour}) & 2     &       & 4     & 3    \\
    & Time to clean segmented output (\si{\hour})  & 2     &       & 0     & 0    \\
    }
The time taken for manual segmentation and for each step in the automatic
segmentation process are also included.
Timings for ASEMI were obtained on a system with an Intel Xeon W-3225 CPU at
\SI{3.70}{\giga\hertz}, \SI{256}{\gibi\byte} DDR4-2666 RAM, and an
NVIDIA GeForce RTX\,2080\,Ti GPU.
Timings for U-Net were obtained on a system with a dual Intel Xeon Platinum 8160
CPU at \SI{2.1}{\giga\hertz}, \SI{1.5}{\tebi\byte} DDR4-2666 RAM, and two
NVIDIA Quadro P6000 GPUs.
It can be readily seen that the training time for ASEMI is already
significantly faster than for U-Net, even though the ASEMI implementation
has only been partially optimised, while U-Net is based on the heavily
optimised Tensorflow.
Unfortunately, the segmentation time is still considerably slower in ASEMI,
due to the current architecture of the implementation, which works off disk.
We believe it is possible to significantly improve this aspect of the
implementation, without losing the advantage of being able to work with
volumes much larger than memory, and without any changes to the algorithm
performance.


\section{Results}

\subsection{Accuracy comparison}

An initial comparison of the different machine learning models is based on
the overall accuracy of the trained model when used to segment parts of the
volume not already seen during training.
Accuracy is given by the intersection-over-union metric, which we compute for
the three models considered, as shown in Table~\ref{tbl:accuracy}.
\inserttabd{tbl:accuracy}{l | c c c | c c c | c c c | c c c}
    {{\bf Intersection-over-union results for individual labels and overall
    accuracy for four specimens.}
    U-Net: Dragonfly implementation;
    RF: ASEMI segmenter with random forest classifier;
    NN: ASEMI segmenter with neural network classifier.}
    {
                  & \multicolumn{3}{c|}{\emph{MHNGr.ET.1023 (dog)}}
                  & \multicolumn{3}{c|}{\emph{MHNGr.ET.1017 (raptor)}}
                  & \multicolumn{3}{c|}{\emph{MHNGr.ET.1456 (ibis)}}
                  & \multicolumn{3}{c}{\emph{MG.2038 (ibis in a jar)}} \\
    \emph{Label}  & \emph{U-Net}  & \emph{RF}     & \emph{NN}
                  & \emph{U-Net}  & \emph{RF}     & \emph{NN}
                  & \emph{U-Net}  & \emph{RF}     & \emph{NN}
                  & \emph{U-Net}  & \emph{RF}     & \emph{NN} \\
    \hline
    Bones         & \bf 92.2\%    & 76.2\%        & 85.5\%
                  &               &               &
                  & \bf 88.7\%    & 86.2\%        & 81.9\%
                  & 93.4\%        & \bf 96.0\%    & 90.5\%  \\
    Teeth         & 43.2\%        & 14.5\%        & \bf 46.6\%
                  &               &               &
                  &               &               &
                  &               &               &         \\
    Feathers      &               &               &
                  &               &               &
                  & \bf 69.3\%    & 61.6\%        & 64.9\%
                  & \bf 67.7\%    & 21.5\%        & 28.7\%  \\
    Soft parts    & \bf 87.4\%    & 70.3\%        & 72.9\%
                  & \bf 77.4\%    & 66.1\%        & 66.4\%
                  &               &               &
                  & \bf 92.8\%    & 58.8\%        & 81.6\%  \\
    Soft powder   &               &               &
                  &               &               &
                  & \bf 74.9\%    & 58.3\%        & 71.3\%
                  &               &               &         \\
    Stomach       &               &               &
                  & \bf 95.9\%    & 84.8\%        & 80.4\%
                  &               &               &
                  &               &               &         \\
    Snails        &               &               &
                  &               &               &
                  & \bf 87.2\%    & 60.0\%        & 54.2\%
                  &               &               &         \\
    Textiles      & \bf 93.1\%    & 83.6\%        & 85.1\%
                  & \bf 96.5\%    & 90.8\%        & 91.4\%
                  & \bf 86.2\%    & 82.7\%        & 82.3\%
                  &               &               &         \\
    Balm textile  &               &               &
                  & \bf 85.3\%    & 81.1\%        & 80.5\%
                  &               &               &
                  &               &               &         \\
    Dense textile &               &               &
                  & \bf 81.0\%    & 66.5\%        & 67.9\%
                  & \bf 67.0\%    & 57.5\%        & 62.8\%
                  & \bf 96.5\%    & 78.0\%        & 78.7\%  \\
    Natron        &               &               &
                  &               &               &
                  & \bf 60.4\%    & 36.9\%        & 22.0\%
                  &               &               &         \\
    Ceramics      &               &               &
                  & \bf 78.8\%    & 67.7\%        & 65.3\%
                  &               &               &
                  &               &               &         \\
    Terracotta    &               &               &
                  &               &               &
                  &               &               &
                  & \bf 99.8\%    & 98.9\%        & 99.3\%  \\
    Cement        &               &               &
                  &               &               &
                  &               &               &
                  & \bf 94.4\%    & 74.8\%        & 78.2\%  \\
    Wood          &               &               &
                  & 84.4\%        & 76.8\%        & \bf 94.8\%
                  &               &               &
                  &               &               &         \\
    Insects       &               &               &
                  &               &               &
                  & \bf 25.4\%    &  4.6\%        &  4.5\%
                  &               &               &         \\
    Powder        &               &               &
                  &               &               &
                  &               &               &
                  & \bf 96.8\%    & 70.9\%        & 71.8\%  \\
    Unlabelled    & \bf 99.2\%    & 97.7\%        & 98.7\%
                  & \bf 97.7\%    & 95.0\%        & 89.9\%
                  & \bf 99.0\%    & 97.1\%        & 98.8\%
                  & \bf 99.4\%    & 97.0\%        & 98.5\%  \\
    \hline
    Overall       & \bf 98.9\%    & 96.5\%        & 97.7\%
                  & \bf 97.2\%    & 94.2\%        & 94.3\%
                  & \bf 97.4\%    & 94.6\%        & 96.8\%
                  & \bf 99.4\%    & 96.0\%        & 97.2\%  \\
    }
The same test set is used for each model, ensuring comparability of results.
It is readily seen that U-Net outperforms ASEMI in almost every label.
Interestingly, the few cases where ASEMI performs better include those
involving labels with a small representation: `teeth' in MHNGr.ET.1023 (dog)
and `wood' in MHNGr.ET.1017 (raptor).
In these cases it seems that the neural network classifier with ASEMI features
is better able to learn an association with less training data.
It would be interesting to see whether this behaviour allows us to achieve
reasonable performance with less manually segmented slices.
Another instance where ASEMI performs better is `bones' in MG.2038 (ibis in
a jar).
In this case we know that there are some errors in the manual segmentation;
it seems that in such cases the random forest classifier with ASEMI features
is able to correct these errors, as we will see in the following analysis.

\subsection{Error analysis}

A quantitative error analysis can be performed by considering
the confusion matrices over the test set predictions, as shown in
Fig~\ref{fig:confusion-matrices} for the random forest classifier.
\insertfigd{fig:confusion-matrices}
    {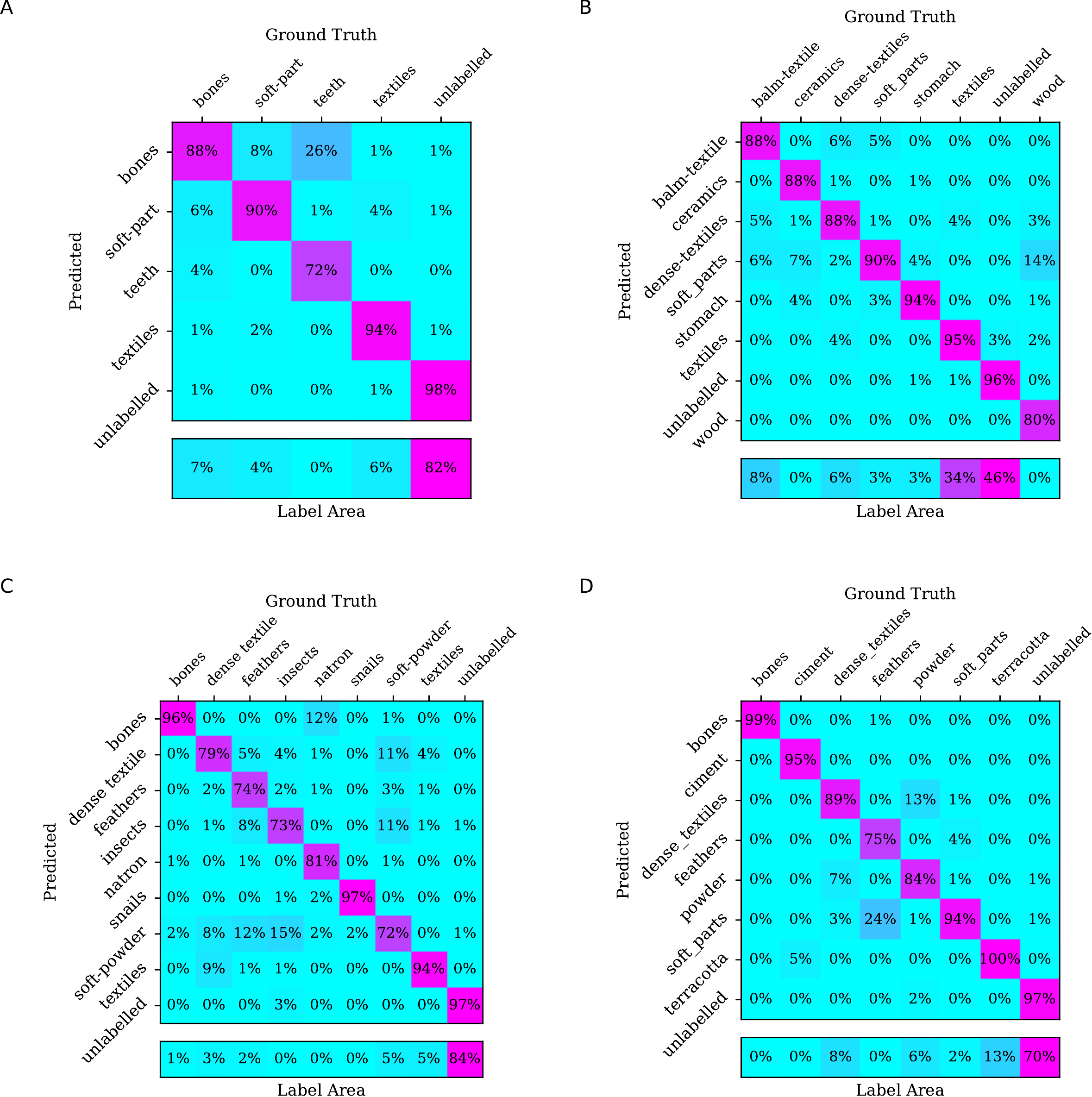}
    {{\bf Confusion matrices for random forest predictions.}
    For each specimen's test set, the confusion matrix is given together
    with the percentage of the number of voxels in the test set occupied by
    a given label.
        A:~MHNGr.ET.1023 (dog),
        B:~MHNGr.ET.1017 (raptor),
        C:~MHNGr.ET.1456 (ibis),
        D:~MG.2038 (ibis in a jar).}
This shows, for every label, what the voxels are classified as.
Numbers on the main diagonal indicate correct classification, while non-zero
off-diagonal values indicate misclassifications.
Fig~\ref{fig:confusion-matrices} also includes, for each label,
the percentage of the number of voxels in the test set occupied by that label.
Some general trends can be observed, particularly that misclassification is
more likely for labels that have a low representation, and between labels
with similar textures.


While the numerical accuracy may be lower, a visual inspection of the
segmentations shows that the overall output from ASEMI is still useful.
In MHNGr.ET.1023 (dog), for example, the largest misclassification is that
26\% of the `teeth' voxels were mislabelled as `bones'.
This is hardly surprising, given the similarity between these materials
and the rather low representation of `teeth' in the samples.
We also see the inverse error, with `bones' being mislabelled as `teeth',
as shown in Fig~\ref{fig:detail_dog}.
\insertfigd{fig:detail_dog}
    {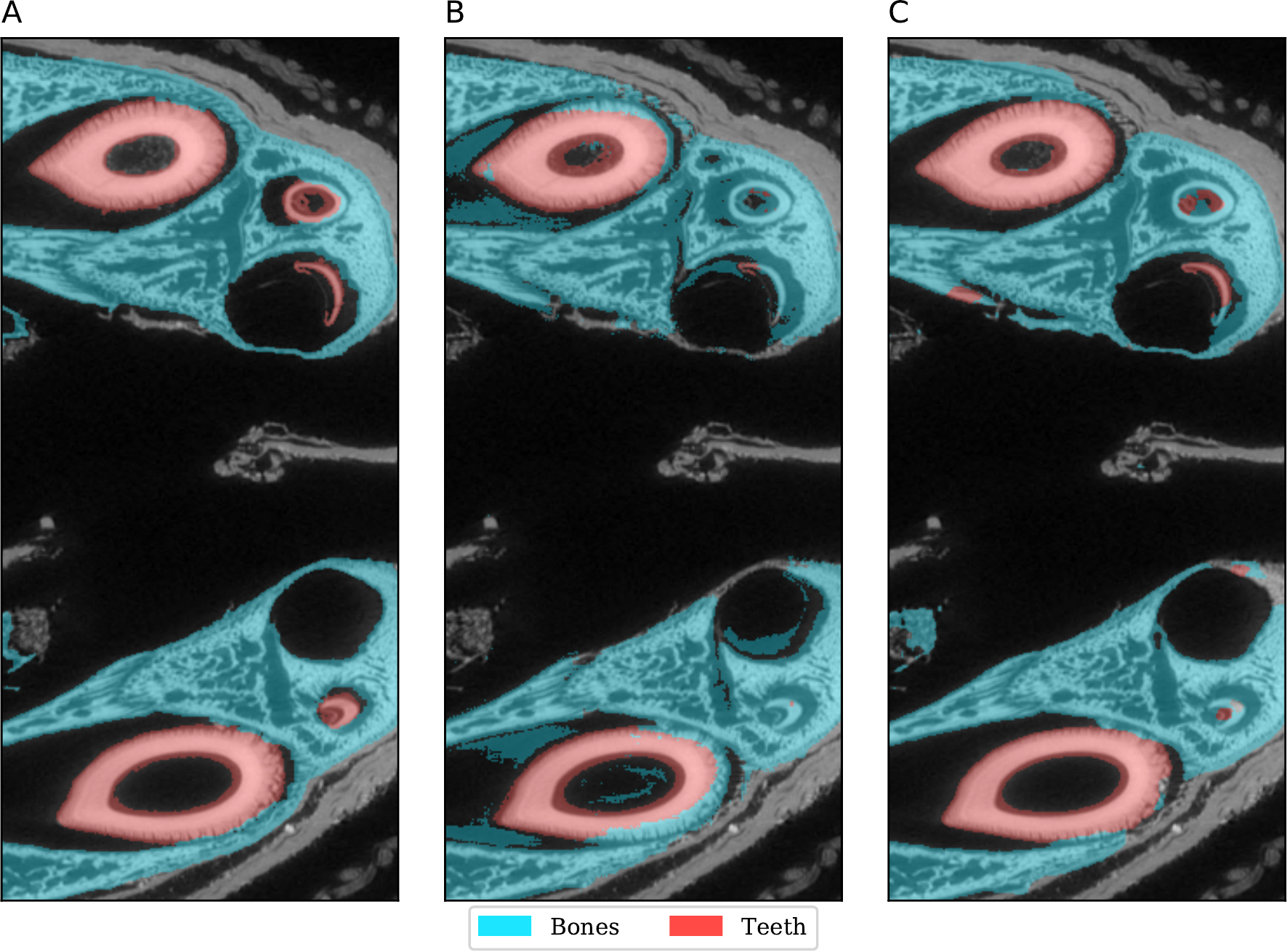}
    {{\bf Segmentation detail in MHNGr.ET.1023 (dog).}
    A:~manual labelling,
    B:~ASEMI output (random forest classifier),
    C:~U-Net output.}
Where teeth and bones are confused, they are very similar in texture and
brightness.
The error is not greater than it is only because the bone developed to the same
density of teeth in small regions of the skull.
Distinguishing between bones and teeth will generally be problematic, due
to similarity in the materials' density and texture.
This is an instance where wider context information is likely to be useful.
It is interesting that even U-Net made similar errors, indicating that the
problem is not a trivial one.


A similar observation can be made in MHNGr.ET.1017 (raptor), where
14\% of `wood' voxels were mislabelled as `soft parts'.
The `wood' voxels have a low representation, while `soft parts' is a rather
diverse collection of voxels, some of which bear similarity to `wood'.
In this specimen we also observe other examples of mislabelling between
similar textures, or where the dividing line is somewhat arbitrary, as shown
in Fig~\ref{fig:detail_raptor}.
\insertfigd{fig:detail_raptor}
    {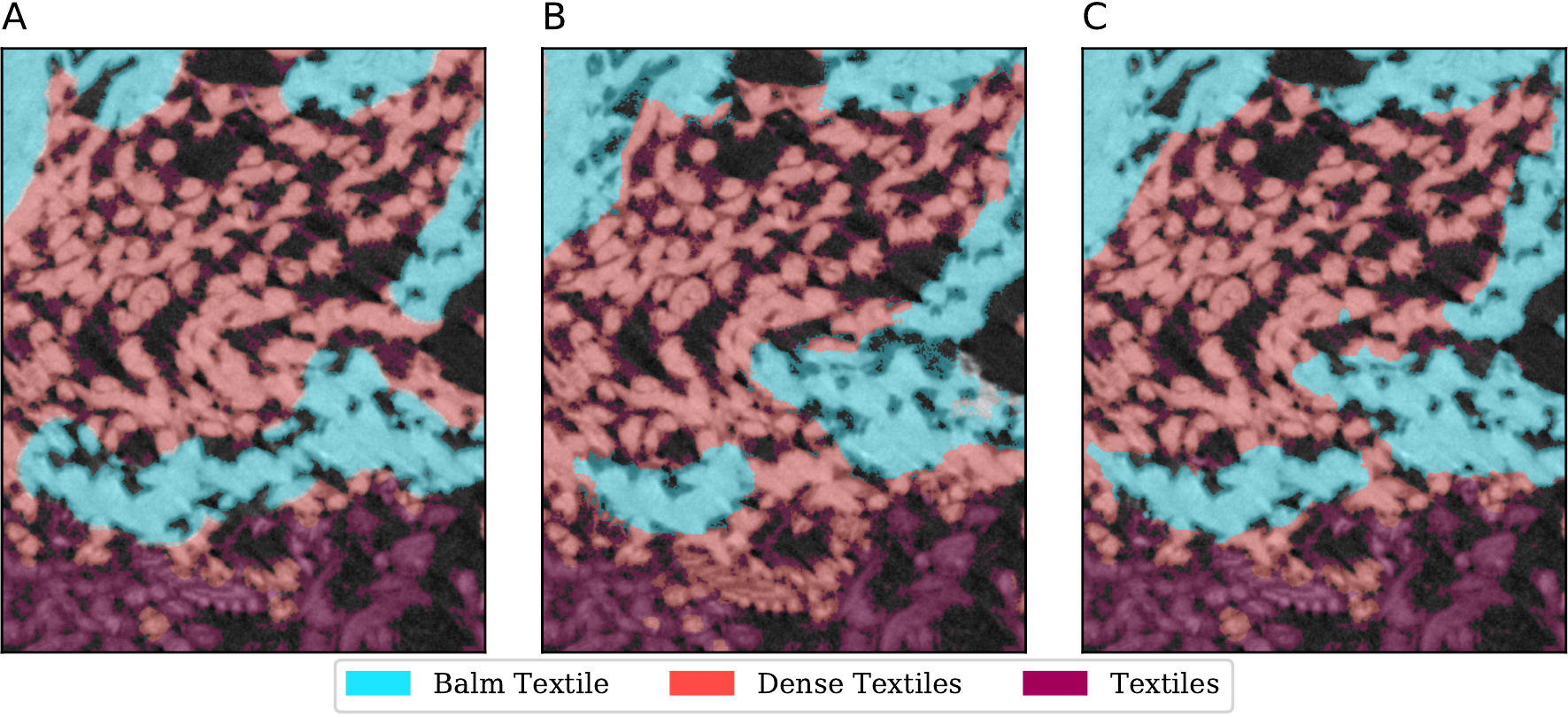}
    {{\bf Segmentation detail in MHNGr.ET.1017 (raptor).}
    A:~manual labelling,
    B:~ASEMI output (random forest classifier),
    C:~U-Net output.}
In this example we can observe the confusion between `textiles', which is
the outer layer of wrapping, `dense textile', which is a denser wrapping,
and `balm-textile', which is textile impregnated with embalming resin.
In particular, `dense textile' and `balm-textile' have similar density
and texture.


In MHNGr.ET.1456 (ibis), we see that 12\% of `natron', which has a low
representation, was misclassified as `bones', which has a similar texture
and voxel intensity.
Diverse labels, such as `soft-powder', are also easily misclassified to labels
that share some texture similarity, such as `dense textiles' or `insects'.
Another interesting error is the confusion of `feathers' with `insects',
as shown in Fig~\ref{fig:detail_ibis}.
\insertfigd{fig:detail_ibis}
    {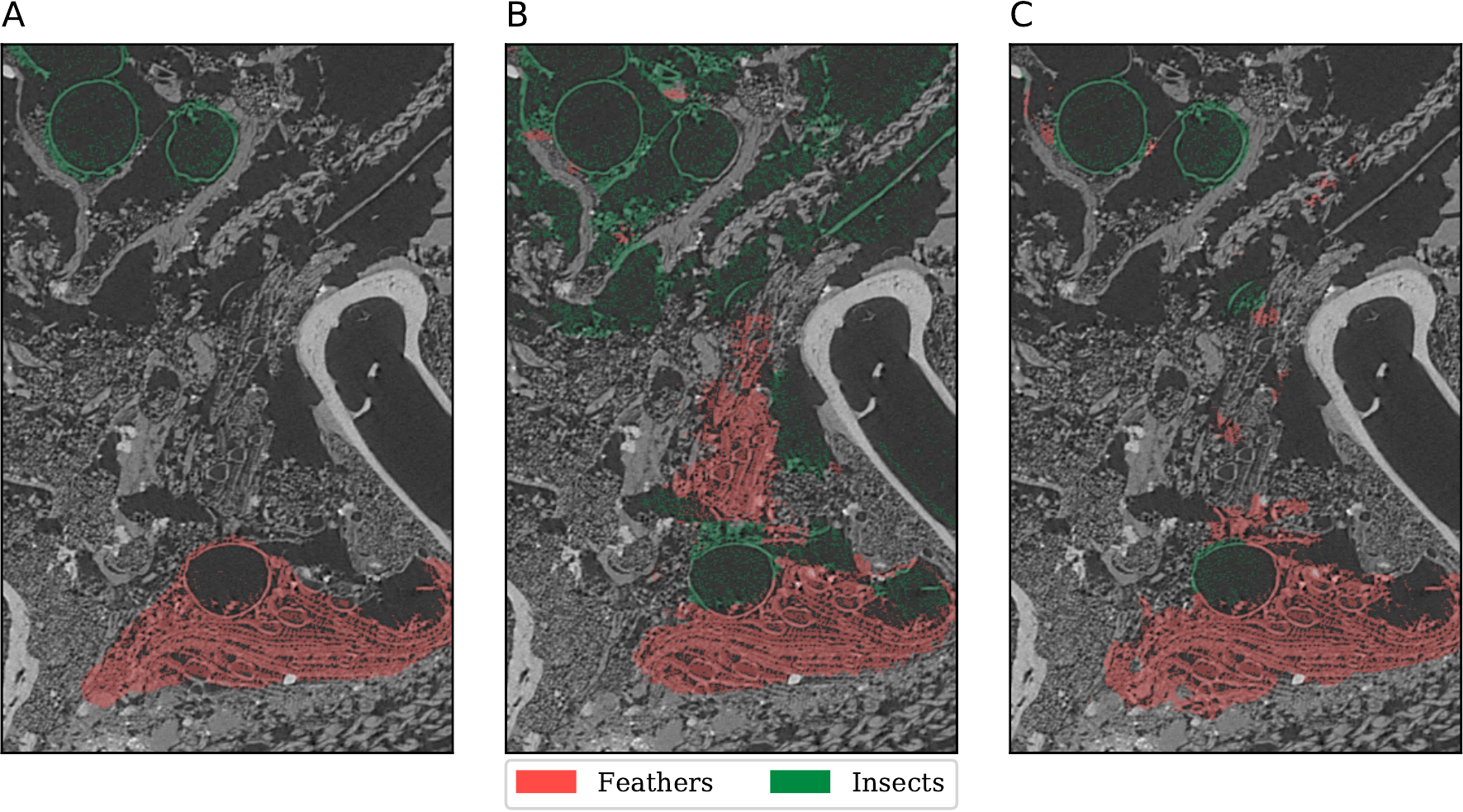}
    {{\bf Segmentation detail in MHNGr.ET.1456 (ibis).}
    A:~manual labelling,
    B:~ASEMI output (random forest classifier),
    C:~U-Net output.}
The insects are generally an empty exoskeleton, so that they appear as
unfilled circular objects in cross-sections of the volumetric image.
Cross-sections of feather stems also appear as unfilled circular objects,
perhaps explaining the confusion.
As we have seen with similar errors in other specimens, these labels may
be distinguished by considering a wider context.
Once again, U-Net also made similar errors, indicating the difficulty of
the task.


Finally, in MG.2038 (ibis in a jar), we see that `feathers', which also have
a low representation, misclassify to `soft\_parts'.
There is also misclassification between `powder' and `dense textiles',
which have boundaries that are not always cleanly defined in the training set.
This is illustrated in Fig~\ref{fig:detail_jar}.
\insertfigd{fig:detail_jar}
    {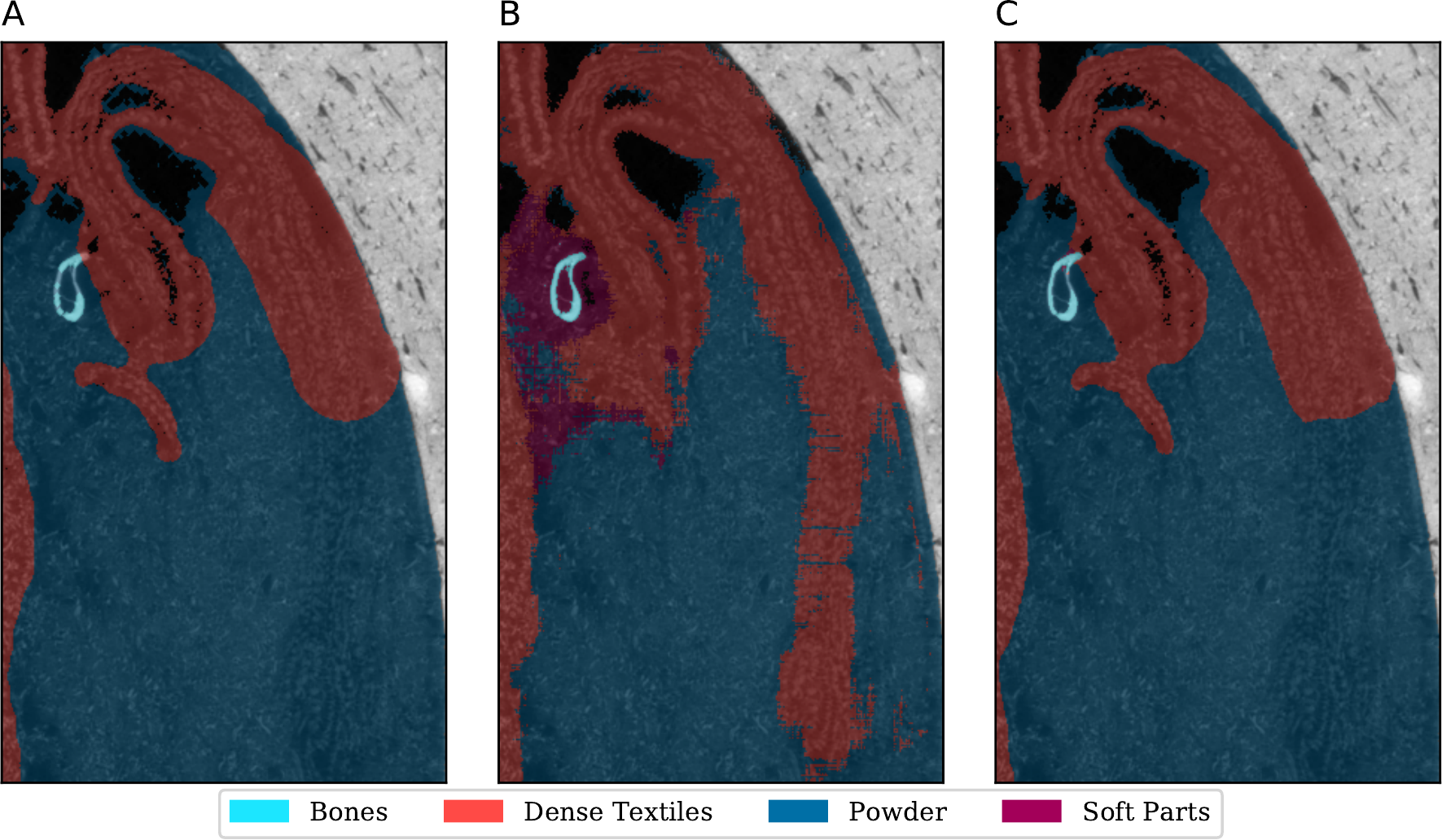}
    {{\bf Segmentation detail in MG.2038 (ibis in a jar).}
    Illustrates uncertainty at label edges for the ASEMI segmenter with random
    forest classifier, some powder being incorrectly labelled as soft parts,
    and correction of manual mislabelling, as compared to U-Net.
    A:~manual labelling,
    B:~ASEMI output (random forest classifier),
    C:~U-Net output.}
It is interesting to see that the reason for this error may be due to a
mislabelling in the manual segmentations.
The figure shows a section of dense textile which is marked as abruptly
ending in the middle of the image, when it seems that it actually extends
further down, which the ASEMI prediction corrects.
Observe how the U-Net output does not perform this correction, indicating
that U-Net is over-fitting to the manual segmentations.
A similar smaller error correction occurs in the bone cross-section.
Fig~\ref{fig:detail_jar} also shows evidence of noisy output,
particularly in edges between labels.
Uncertainty at the edges can be easily fixed using morphological operations;
an alternative is to use probabilistic filters, such as Markov random fields.

\subsection{3D rendering of labelled output}

Finally, 3D labelled reconstructions of three specimens can be seen in
Figs~\ref{fig:3d_dog}--\ref{fig:3d_ibis_jar}.
For MHNGr.ET.1023 (dog), in Fig~\ref{fig:3d_dog}, the ASEMI segmentation
with the neural network classifier is used to render the external wrappings
and the soft tissue.
\insertfigd{fig:3d_dog}
    {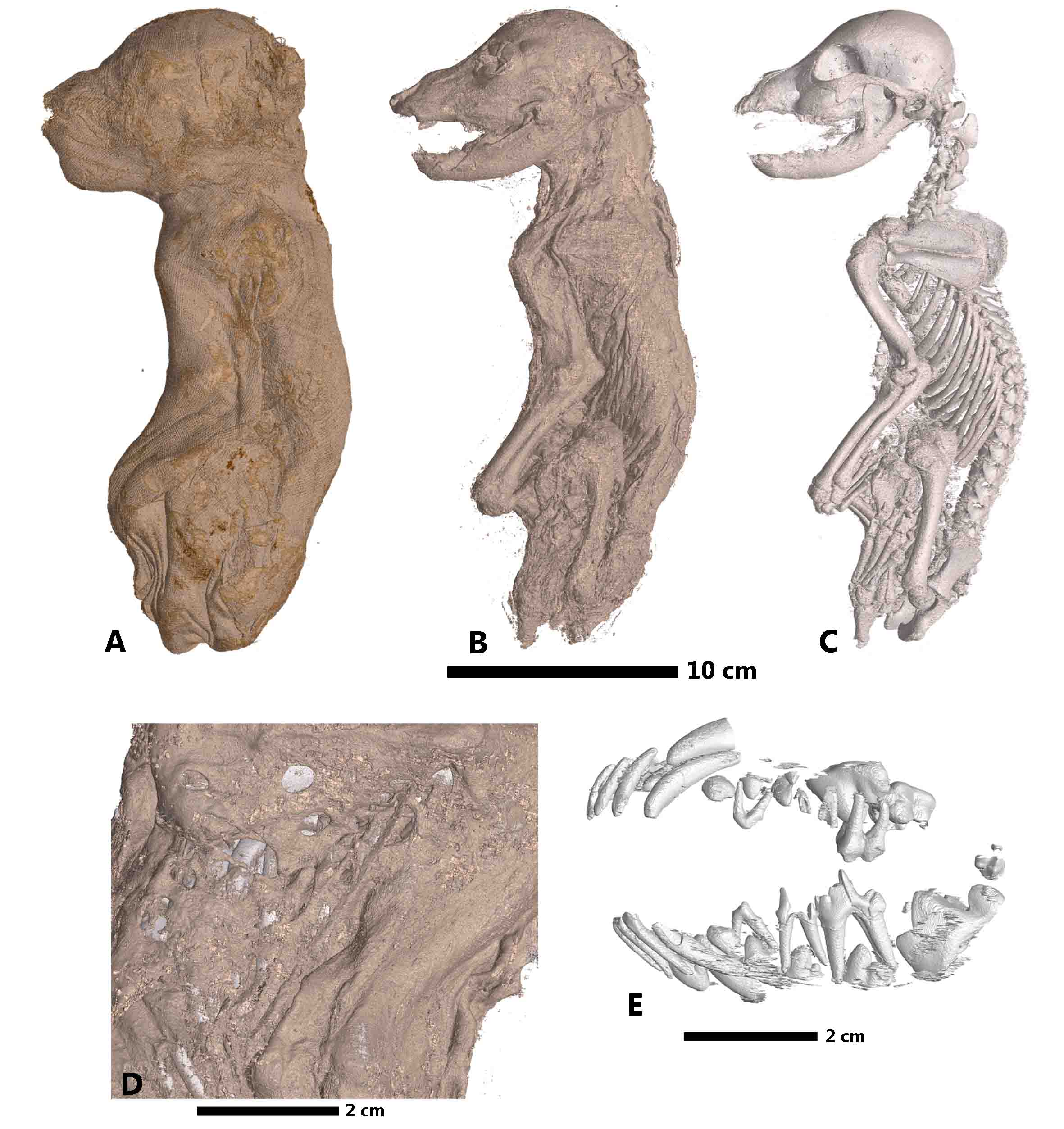}
    {{\bf Rendering of the mummy MHNGr.ET.1023 (dog).}
        A:~external aspect of the remaining bandages (segmentation ASEMI NN),
        B:~rendering of the preserved soft parts of the puppy (segmentation
           ASEMI NN),
        C:~rendering of the bones (segmentation U-Net),
        D:~detail on the soft parts showing its weak preservation state
           through holes certainly caused by pests or putrefaction
           (segmentation U-Net),
        E:~rendering of the decidual and permanent dentition of puppy
           showing its young age (segmentation U-Net).}
Renderings of the bones and teeth use the U-Net segmentation.
A detailed image of the soft parts, using the U-Net segmentation, shows
the weak preservation state of the mummy, with holes caused by pests or
putrefaction.
The U-Net segmentation of the teeth (Fig~\ref{fig:3d_dog}, E) allows us to
estimate its age between six and twelve weeks old.
These different segmentations permit us to know that this mummy was made
from the decaying corpse of a relatively young puppy.
Reconstruction of MHNGr.ET.1456 (ibis) is shown in Fig~\ref{fig:3d_ibis}.
\insertfigd{fig:3d_ibis}
    {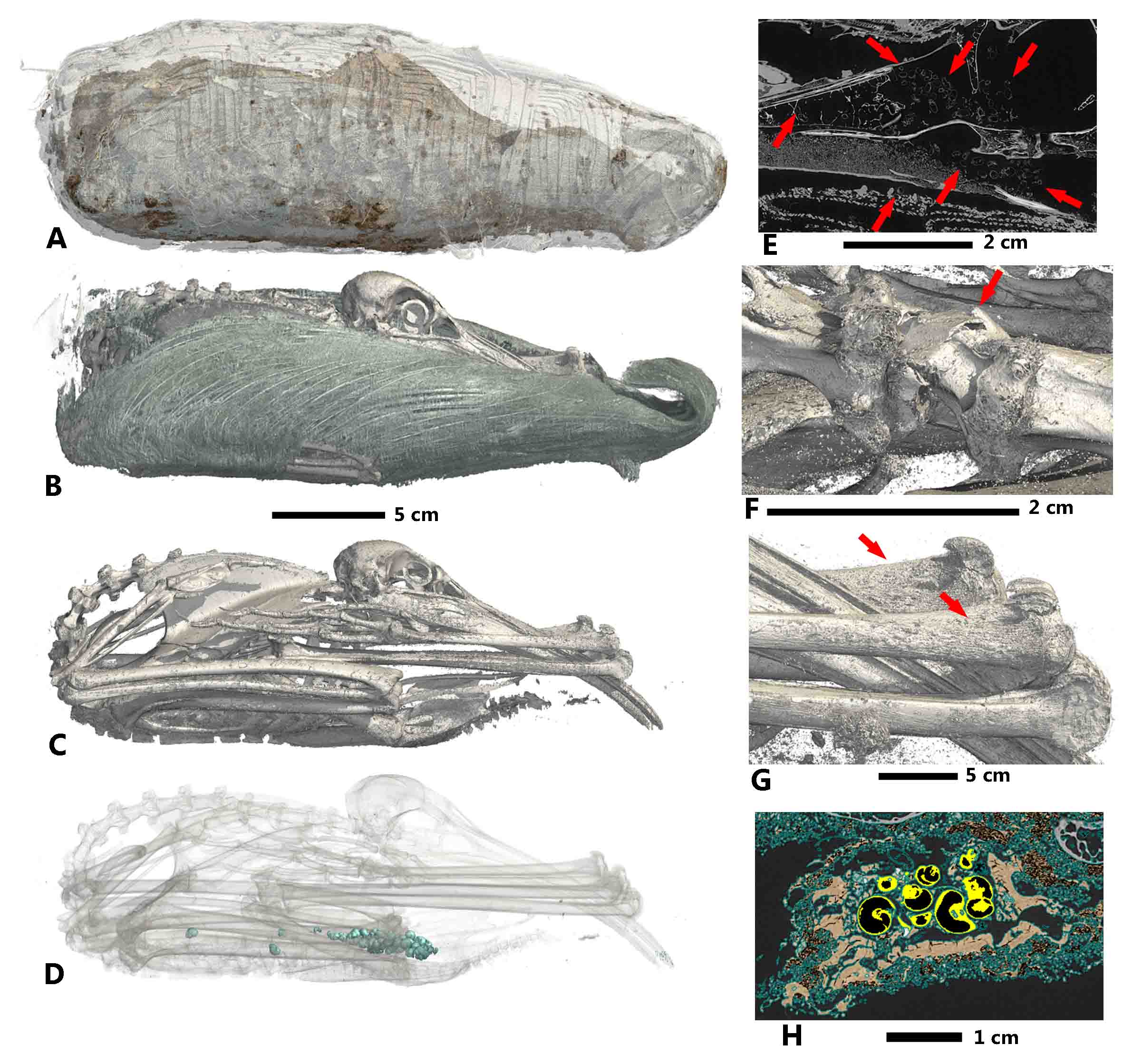}
    {{\bf Mummy MHNGr.ET.1456 (ibis).}
        A:~general 3D rendering (segmentation ASEMI NN),
        B:~3D rendering of feathers of the ibis (segmentation U-Net),
        C:~3D rendering of the skeleton (segmentation U-Net),
        D:~3D rendering of the gastropods (segmentation U-Net),
        E:~virtual slide showing the pest infestation inside the mummy,
        F:~3D rendering of the broken vertebra and the twist of the neck
           (ASEMI NN),
        G:~3D rendering of the osteoporosis on the forelimbs (ASEMI NN),
        H:~virtual slide of the stomach (orange highlight) and the gastropods
           (yellow highlight) (segmentation U-Net).}
The ASEMI segmenter with the neural network classifier is used for the
general 3D render.
The U-Net segmentations was used to image the feathers and the skeleton,
as well as the gastropods present inside the stomach of the bird.
Using the ASEMI segmenter with the neural network classifier, it was possible
to render the neck of the bird, highlighting twisting of the neck and a
broken vertebra.
The ASEMI segmenter also renders detail of the forelimbs showing osteoporosis.
All the segmentations raise questions about the health of the ibis inside
the mummy.
This can be illustrated by the fact that the cervical fracture may correspond
to the cause of the death of the bird.
For MG.2038 (ibis in a jar), reconstructions are shown in
Fig~\ref{fig:3d_ibis_jar}.
\insertfigd{fig:3d_ibis_jar}
    {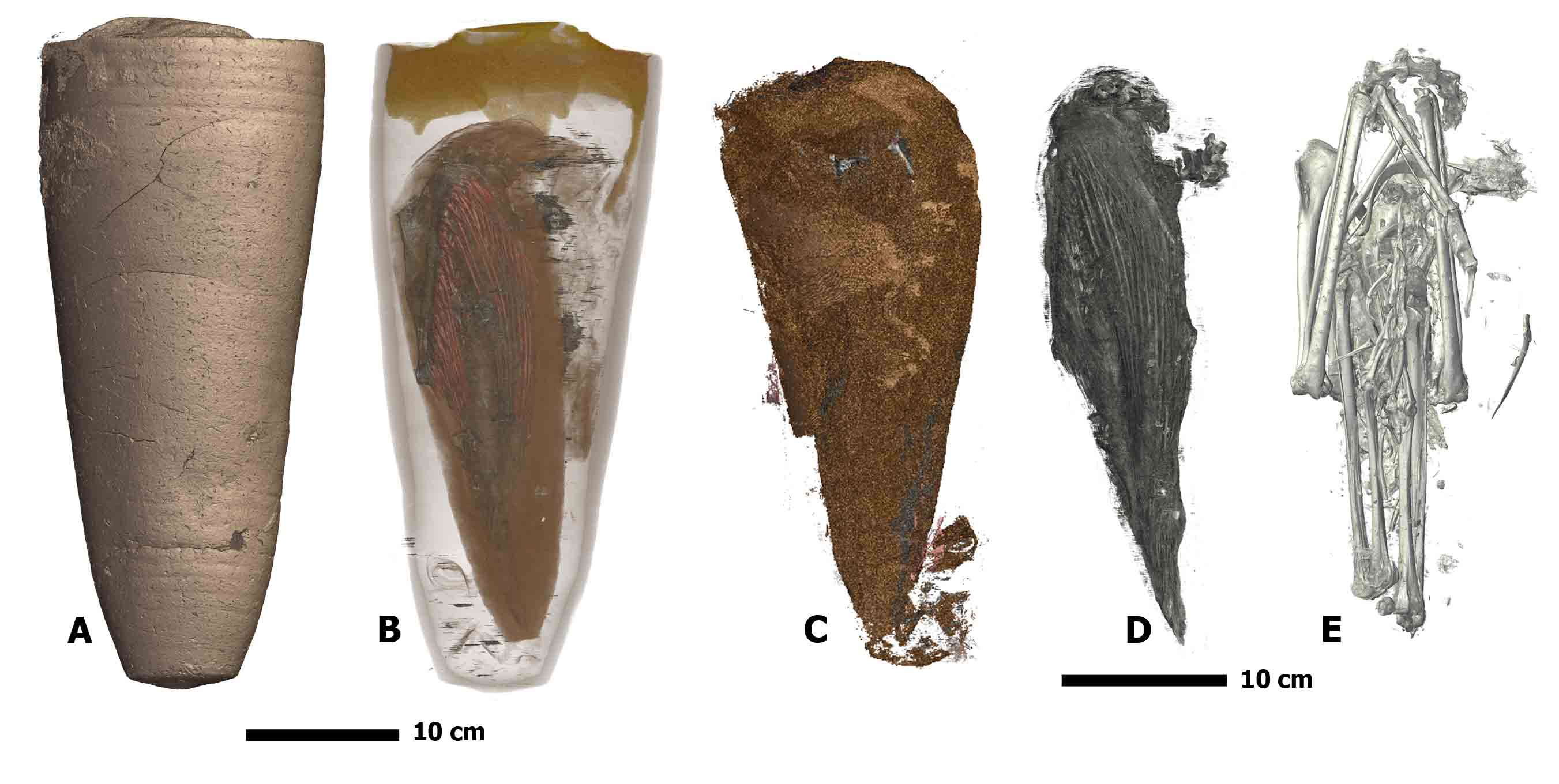}
    {{\bf 3D rendering of the mummy MG.2038 (ibis in a jar).}
        A:~general rendering of the external aspect of the mummy (segmentation
           ASEMI RF),
        B:~internal content of the mummy made visible thanks to the
           transparency of the sealed jar (segmentation U-Net),
        C:~focus on the textile surrounding the animal (segmentation ASEMI
           RF), focus on the ibis inside the mummy
        D:~showing its soft part (segmentation ASEMI NN),
        E:~showing the ibis bones (segmentation ASEMI RF).}
The ASEMI segmenter with random forest classifier is used for the render of
the container.
The U-Net segmenter is used to show the interior, by rendering the jar
itself transparently.
The ASEMI segmenter is used to show the textiles, soft parts, and bones.
These different images have brought knowledge on the ibis mummy but also on
the jar that contained it.
The jar was made on a wheel rotating clockwise and was sealed with a cover
glued with a type of mortar.
Tomographic data from mummy MHNGr.ET.1017 have shown that the mummy does
not contain a complete bird raptor but only its head stuck on the wrapped
mummy and fixed through a vegetal stem as well as a very juvenile sea bird
in the centre of the fabrics.
However, due to the focusing on the internal part of the mummy, the different
segmentations show mainly different type of tissues soaked with balm, which
make the rendering very confusing for non-specialist eyes.



\section{Conclusions}

In the Automated SEgmentation of Microtomography Imaging (ASEMI) project we
have developed a tool to automatically segment volumetric images of Egyptian
mummies, obtained using Propagation Phase Contrast Synchrotron Microtomography
(PPC-SRμCT).
In contrast to emerging commercial solutions, our tool uses simple 3D features
and classical machine learning models, for a much lower theoretical complexity.
Indicative results were given for four specimens, showing similar overall
accuracy when compared with manual segmentations (94--98\% as compared with
97--99\% for deep learning).
A qualitative analysis was also given, showing that our results are close
in terms of usability to those from deep learning.

While this work demonstrates the feasibility of using machine learning for the
3D segmentation of large volumes, a number of further advances are
necessary for an operational environment.
As we have shown, the output segmentations require some postprocessing to
smoothen noisy edges.
Preliminary work using a Markov Random Field gave promising results, and we
plan to implement this in a scalable way.
Current 3D deep learning segmentation methods do not scale well to large
volumes.
If our complexity reduction technique can be applied to established deep
learning models, it would become feasible to use deep learning with a full
3D context.
This would allow us to get the accuracy we have observed with deep learning
models without the discontinuities that exist with current 2D-context
approaches.
One of the limitations of the current approach is the need for manual
segmentation of a small set of slices for training purposes.
Extending our segmenter to use an incremental learning approach would allow the
iterative improvement of a learned model as further specimens are segmented.
This would enable the use of a model trained on one or more specimens to
segment a completely new specimen, with reduced user input.
Such an extension is necessary for the automatic segmentation of collections
of related mummies, which is currently not feasible.

\FloatBarrier


\section{Acknowledgment}
This project has received funding from the ATTRACT project funded by the EC
under Grant Agreement 777222.
Authors thank Alessandro Mirone for the initial GPU implementation of the
3D histogram algorithm, and Andy Götz for project management support.
We would like to acknowledge the Museum d’Histoire Naturelle de Grenoble
(France), in particular Philippe Candegabe, for the opportunity to scan
the mummies MHNGr.ET.1017, MHNGr.ET.1023, MHNGr.ET.1456.
We also want to thank the Musée de Grenoble (France) in the person of
Valérie Huss for providing access to the sealed jar MG.2038.


\bibliography{IEEEabrv,references}

\begin{thebibliography}{10}

\bibitem{Berruyer2020}
Berruyer C, Porcier SM, Tafforeau P.
\newblock Synchrotron {\textquotedblleft}virtual
  archaeozoology{\textquotedblright} reveals how Ancient Egyptians prepared a
  decaying crocodile cadaver for mummification.
\newblock {PLOS} {ONE}. 2020;15(2):e0229140.
\newblock doi:{10.1371/journal.pone.0229140}.

\bibitem{Porcier2019}
Porcier SM, Berruyer C, Pasquali S, Ikram S, Berthet D, Tafforeau P.
\newblock Wild Crocodiles Hunted to Make Mummies in Roman Egypt: Evidence from
  Synchrotron Imaging.
\newblock Journal of Archaeological Science. 2019;110:105009.
\newblock doi:{10.1016/j.jas.2019.105009}.

\bibitem{vrooman2006}
Vrooman HA, Cocosco CA, Stokking R, Ikram MA, Vernooij MW, Breteler MM, et~al.
\newblock {kNN}-based multi-spectral {MRI} brain tissue classification: manual
  training versus automated atlas-based training.
\newblock In: Medical Imaging 2006: Image Processing. vol. 6144. International
  Society for Optics and Photonics; 2006. p. 61443L.

\bibitem{cremers2007}
Cremers D, Rousson M, Deriche R.
\newblock A review of statistical approaches to level set segmentation:
  integrating color, texture, motion and shape.
\newblock International journal of computer vision. 2007;72(2):195--215.

\bibitem{zhang2017}
Zhang L, Wang L, Yang B, Chen Z, Zhou J, Han Y, et~al.
\newblock Three dimensional segmentation for cement microtomography images
  using self-organizing map and neighborhood features.
\newblock In: 2017 IEEE Symposium Series on Computational Intelligence (SSCI).
  IEEE; 2017. p. 1--8.

\bibitem{loeffler2018}
Loeffler CM, Qiu Y, Martin B, Heard W, Williams B, Nie X.
\newblock Detection and segmentation of mechanical damage in concrete with
  {X-ray} microtomography.
\newblock Materials Characterization. 2018;142:515--522.

\bibitem{chandar2016}
Chandar KP, Satyasavithri T.
\newblock Segmentation and {3D} visualization of pelvic bone from {CT} scan
  images.
\newblock In: 2016 IEEE 6th International Conference on Advanced Computing
  (IACC). IEEE; 2016. p. 430--433.

\bibitem{ma2010}
Ma Z, Tavares JMR, Jorge RN, Mascarenhas T.
\newblock A review of algorithms for medical image segmentation and their
  applications to the female pelvic cavity.
\newblock Computer Methods in Biomechanics and Biomedical Engineering.
  2010;13(2):235--246.

\bibitem{majd2010}
Majd EM, Sheikh UU, Abu-Bakar S.
\newblock Automatic segmentation of abdominal aortic aneurysm in computed
  tomography images using spatial fuzzy C-means.
\newblock In: 2010 Sixth International Conference on Signal-Image Technology
  and Internet Based Systems. IEEE; 2010. p. 170--175.

\bibitem{haberl2018}
Haberl MG, Churas C, Tindall L, Boassa D, Phan S, Bushong EA, et~al.
\newblock {CDeep3M} -- Plug-and-Play cloud-based deep learning for image
  segmentation.
\newblock Nature methods. 2018;15(9):677--680.

\bibitem{Ronneberger2015}
Ronneberger O, Fischer P, Brox T.
\newblock U-NET: Convolutional Networks for Biomedical Image Segmentation.
\newblock In: Medical Image Computing and Computer-assisted Intervention
  (miccai). vol. 9351 of Lncs. Springer; 2015. p. 234--241.
\newblock Available from:
  \url{http://lmb.informatik.uni-freiburg.de/Publications/2015/RFB15a}.

\bibitem{badrinarayanan2017}
Badrinarayanan V, Kendall A, Cipolla R.
\newblock {SegNet}: A deep convolutional encoder-decoder architecture for image
  segmentation.
\newblock IEEE transactions on pattern analysis and machine intelligence.
  2017;39(12):2481--2495.

\bibitem{qi2017}
Qi CR, Su H, Mo K, Guibas LJ.
\newblock {PointNet}: Deep learning on point sets for {3D} classification and
  segmentation.
\newblock In: Proceedings of the IEEE conference on computer vision and pattern
  recognition; 2017. p. 652--660.

\bibitem{wang2018}
Wang W, Yu R, Huang Q, Neumann U.
\newblock {SGPN}: Similarity group proposal network for {3D} point cloud
  instance segmentation.
\newblock In: Proceedings of the IEEE conference on computer vision and pattern
  recognition; 2018. p. 2569--2578.

\bibitem{Kavur2020}
Kavur AE, Gezer NS, Barış M, Aslan S, Conze PH, Groza V, et~al.
\newblock {CHAOS} Challenge -- combined {(CT-MR)} healthy abdominal organ
  segmentation.
\newblock Medical Image Analysis. 2021;69:101950.
\newblock doi:{10.1016/j.media.2020.101950}.

\bibitem{Hati2020}
Hati A, Bustreo M, Sona D, Murino V, Bue AD.
\newblock Weakly Supervised Geodesic Segmentation of Egyptian Mummy {CT} Scans.
\newblock CoRR. 2020;abs/2004.08270.

\bibitem{OMahoney2019}
O'Mahoney T, Mcknight L, Lowe T, Mednikova M, Dunn J.
\newblock A machine learning based approach to the segmentation of micro {CT}
  data in archaeological and evolutionary sciences.
\newblock bioRxiv. 2020;doi:{10.1101/859983}.

\bibitem{Friedman2012}
Friedman SN, Nguyen N, Nelson AJ, Granton PV, MacDonald DB, Hibbert R, et~al.
\newblock Computed Tomography ({CT}) Bone Segmentation of an Ancient Egyptian
  Mummy A Comparison of Automated and Semiautomated Threshold and Dual-Energy
  Techniques.
\newblock Journal of Computer Assisted Tomography. 2012;36(5):616--622.
\newblock doi:{10.1097/rct.0b013e31826739f5}.

\bibitem{dragonfly}
{Object Research Systems (ORS)}. Dragonfly; 2020.
\newblock Available from:
  \url{https://www.theobjects.com/dragonfly/index.html}.

\bibitem{asemi-datasets}
Automated segmentation of microtomography imaging of Egyptian mummies; 2021.
\newblock Available from: \url{http://paleo.esrf.eu/}.

\bibitem{asemi-segmenter}
{ASEMI} Segmenter; 2021.
\newblock Available from: \url{https://github.com/um-dsrg/ASEMI-segmenter}.

\bibitem{Ojala1996}
Ojala T, Pietik{\"{a}}inen M, Harwood D.
\newblock A Comparative Study of Texture Measures with Classification Based on
  Featured Distributions.
\newblock Pattern Recognition. 1996;29(1):51--59.
\newblock doi:{10.1016/0031-3203(95)00067-4}.

\bibitem{Zhao2007}
Zhao G, Pietikainen M.
\newblock Dynamic Texture Recognition Using Local Binary Patterns with an
  Application to Facial Expressions.
\newblock {IEEE} Transactions on Pattern Analysis and Machine Intelligence.
  2007;29(6):915--928.
\newblock doi:{10.1109/tpami.2007.1110}.

\bibitem{Abbasi2017}
Abbasi S, Tajeripour F.
\newblock Detection of brain tumor in 3D {MRI} images using local binary
  patterns and histogram orientation gradient.
\newblock Neurocomputing. 2017;219:526--535.
\newblock doi:{10.1016/j.neucom.2016.09.051}.

\bibitem{barkan2013}
{Barkan} O, {Weill} J, {Wolf} L, {Aronowitz} H.
\newblock Fast High Dimensional Vector Multiplication Face Recognition.
\newblock In: 2013 IEEE International Conference on Computer Vision; 2013. p.
  1960--1967.

\bibitem{lowe1999}
Lowe DG.
\newblock Object Recognition from Local Scale-invariant Features.
\newblock In: Proceedings of the Seventh {IEEE} International Conference on
  Computer Vision. vol.~2. {IEEE}; 1999. p. 1150--1157.

\bibitem{bay2008}
Bay H, Ess A, Tuytelaars T, {Van Gool} L.
\newblock Speeded-Up Robust Features (SURF).
\newblock Computer Vision and Image Understanding. 2008;110(3):346 -- 359.
\newblock doi:{https://doi.org/10.1016/j.cviu.2007.09.014}.

\bibitem{ho1998}
{Tin Kam Ho}.
\newblock The random subspace method for constructing decision forests.
\newblock IEEE Transactions on Pattern Analysis and Machine Intelligence.
  1998;20(8):832--844.
\newblock doi:{10.1109/34.709601}.

\bibitem{scikit-learn}
Pedregosa F, Varoquaux G, Gramfort A, Michel V, Thirion B, Grisel O, et~al.
\newblock Scikit-learn: Machine Learning in {P}ython.
\newblock Journal of Machine Learning Research. 2011;12:2825--2830.

\bibitem{bishop1995}
Bishop CM.
\newblock Neural Networks for Pattern Recognition.
\newblock USA: Oxford University Press, Inc.; 1995.

\bibitem{tensorflow2015-whitepaper}
Abadi M, Agarwal A, Barham P, Brevdo E, Chen Z, Citro C, et~al.. {TensorFlow}:
  Large-Scale Machine Learning on Heterogeneous Systems; 2015.
\newblock Available from: \url{https://www.tensorflow.org/}.

\bibitem{cristianini2000}
Cristianini N, Shawe-Taylor J.
\newblock An Introduction to Support Vector Machines and Other Kernel-based
  Learning Methods.
\newblock Cambridge University Press; 2000.

\bibitem{cuda-pg-10-0}
{NVIDIA} {CUDA} {C} Programming Guide; 2018.

\bibitem{Cicek2016}
{\c{C}}i{\c{c}}ek {\"O}, Abdulkadir A, Lienkamp SS, Brox T, Ronneberger O.
\newblock 3D U-Net: Learning Dense Volumetric Segmentation from Sparse
  Annotation.
\newblock In: Ourselin S, Wells WS, Sabuncu MR, Unal G, Joskowicz L, editors.
  Medical Image Computing and Computer-Assisted Intervention (MICCAI). vol.
  9901 of LNCS. Springer; 2016. p. 424--432.
\newblock Available from:
  \url{http://lmb.informatik.uni-freiburg.de/Publications/2016/CABR16}.

\bibitem{Mirone2014}
Mirone A, Brun E, Gouillart E, Tafforeau P, Kieffer J.
\newblock The {PyHST}2 hybrid distributed code for high speed tomographic
  reconstruction with iterative reconstruction and a priori knowledge
  capabilities.
\newblock Nuclear Instruments and Methods in Physics Research Section B: Beam
  Interactions with Materials and Atoms. 2014;324:41--48.
\newblock doi:{10.1016/j.nimb.2013.09.030}.

\bibitem{Paganin2002}
Paganin D, Mayo SC, Gureyev TE, Miller PR, Wilkins SW.
\newblock Simultaneous phase and amplitude extraction from a single defocused
  image of a homogeneous object.
\newblock Journal of Microscopy. 2002;206(1):33--40.
\newblock doi:{10.1046/j.1365-2818.2002.01010.x}.

\end{thebibliography}


\end{document}